\documentclass[runningheads]{llncs}

\usepackage[year=2026]{eccv}

\usepackage{eccvabbrv}

\usepackage{graphicx}
\usepackage{booktabs}

\usepackage[accsupp]{axessibility}  
\usepackage{multirow}

\usepackage{hyperref}

\usepackage{orcidlink}

\newcommand{\mat}[1]{\mathbf{#1}}
\newcommand{\vect}[1]{\boldsymbol{#1}}
\newcommand{\ten}[1]{\mathcal{#1}}

\makeatletter
\renewcommand{\@fnsymbol}[1]{%
    \ifcase#1\or $\ast$\or $\dagger$\or $\ddagger$\or $\S$\or $\P$\or $\|$\or $\#$\else\@ctrerr\fi}
\makeatother

\begin{document}


\title{RaysUp: Ultra-light Universal Feature Upsampling via Geometry-Aware Ray Representation}

\titlerunning{RaysUp: Ultra-light Universal Feature Upsampling}

\author{Yuchuan Ding\thanks{Equal contributions (Co-first authors).}\orcidlink{0009-0004-0113-5166} \and
Linfei Li$^*$\orcidlink{0009-0001-7210-5261} \and
Lin Zhang\thanks{Corresponding author.}\orcidlink{0000-0002-4360-5523}  \and
Ying Shen\orcidlink{0000-0002-2966-7955}} 

\authorrunning{Ding et al.}

\institute{School of Computer Science and Technology, Tongji University, Shanghai, China \\
\email{\{2534061,cslinfeili,cslinzhang,yingshen\}@tongji.edu.cn}}

\maketitle

\begin{abstract}
Pre-trained Vision Foundation Models (VFMs) have become central to modern computer vision due to their powerful semantic representations and strong generalization ability. However, their patchified or pooled outputs are inherently low-resolution, limiting their effectiveness in tasks requiring fine-grained, pixel-level reasoning. Existing feature upsampling approaches either degrade semantic fidelity or rely on VFM-specific retraining and heavy architectures, hindering efficiency and scalability. To address these challenges, we propose RaysUp, an ultra-lightweight, task-agnostic, and VFM-agnostic feature upsampling framework that reconstructs high-resolution feature maps at arbitrary resolutions. Unlike conventional 2D interpolation or attention-based schemes, RaysUp lifts feature reconstruction into a geometry-aware ray domain. Specifically, we introduce a Spatially Decoupled Guidance Encoder for direction-aware guidance encoding, an Any-Resolution Cross-Attention mechanism for resolution-flexible reconstruction, and a novel Ray Positional Encoding (RayPE) that injects implicit 3D geometric priors via 6D Plücker ray coordinates. Finally, A Geometry-Aware Neighborhood Attention module further ensures content-adaptive bilateral aggregation while preserving geometric consistency. Extensive experiments across diverse dense prediction tasks demonstrate that RaysUp achieves state-of-the-art performance while using only 16\% of the parameters of AnyUp and delivering approximately 7$\times$ faster inference. These results highlight a substantially improved accuracy–efficiency trade-off and establish RaysUp as a practical and scalable solution for universal feature upsampling. Code is available at \href{https://github.com/MAP-RaysUp/RaysUp}{https://github.com/MAP-RaysUp/RaysUp}.

\end{abstract}

\section{Introduction}
\label{sec:intro}

\begin{figure*}[t!]
    \centering
    \includegraphics[width=\linewidth]{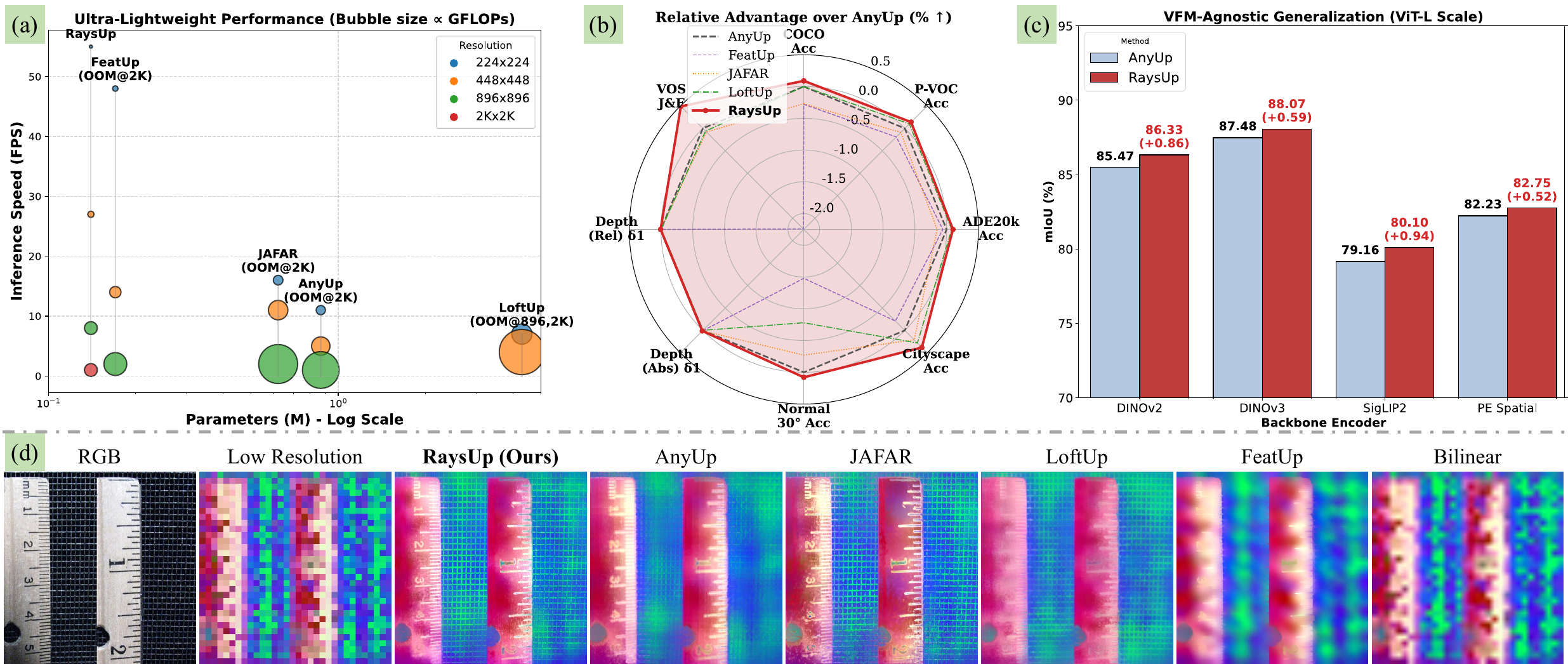}
    \caption{We propose RaysUp, an (a) ultra-lightweight, (b) task-agnostic, and (c) VFM-agnostic upsampling framework, capable of upsampling backbone features to arbitrary resolutions while preserving (d) high semantic fidelity and geometric consistency.}
    \label{fig:teaser}
    \vspace{-15px}
\end{figure*}

Pre-trained Vision Foundation Models (VFMs), such as the DINO family \cite{caron2021dino, Oquab24dinov2, simeoni2025dinov3}, the CLIP series \cite{Radford21CLIP, xu2023metaclip, chuang2025metaclip2}, the SigLIP series \cite{Zhai23siglip, tschannen2025siglip2}, PE Spatial \cite{bolya2025pespatial}, and MAE \cite{he2022mae}, have emerged as central components in modern computer vision due to their large-scale pretraining on diverse datasets, which endows them with high-level semantic representations and strong generalization capability. These models are capable of extracting transferable and task-agnostic visual features, making them highly suitable as either frozen or fine-tuned backbone networks. Consequently, they have been extensively adopted across a wide range of downstream applications, including depth estimation \cite{cvpr2024depthanything, yang2024da2, lin2026da3, chen2025videodepth}, open-vocabulary semantic segmentation \cite{jose2025dinov2txt, shin2024pixelclip, wysoczanska2024clipdinoiser, barsellotti2025talking}, 3D semantic field reconstruction \cite{kerr2023lerf, qin2024langsplat, jun2025dr,li2024gs3lam}, and the evaluation of generative models \cite{asim2025met3r, xie2025mvgbench}.

However, due to patchification or aggressive pooling operations in VFMs, their output features often have low spatial resolution, limiting their applicability in tasks that require fine-grained, pixel-level understanding. Traditional upsampling methods \cite{duchon1979lanczos, mckinley1998cubic}, such as bilinear interpolation, are computationally efficient but often result in the loss of semantic information. More recently, learned upsampling methods, such as LiFT \cite{eccv2024lift}, FeatUp \cite{fu2024featup}, LoftUp \cite{iccv2025loftup}, and JAFAR \cite{couairon2025jafar}, have demonstrated task-agnostic generalization. However, they require retraining for different VFMs or additional optimization at inference time, which restricts their broader applicability. A more recent approach, AnyUp \cite{wimmer2026anyup}, introduces an encoder-agnostic upsampling framework that generalizes across different VFMs, but its reliance on a heavy network architecture makes it difficult to balance generalization capability with inference efficiency.

To address the aforementioned challenges, we propose RaysUp, an ultra-lightweight, task-agnostic, and encoder-agnostic universal feature upsampling framework that enables feature maps extracted from arbitrary Vision Foundation Models to be reconstructed at any target resolution.

From a methodological perspective, RaysUp builds upon the classical Joint Bilateral Upsampling (JBU) \cite{KopfCLU07JBU} paradigm while introducing structural generalizations. Specifically, instead of directly employing the raw RGB image as guidance, RaysUp adopts a lightweight Spatially Decoupled Guidance Encoder to generate guidance features. This encoder captures multi-scale spatial semantics and provides more informative structural priors than pixel-level color differences. Conditioned on these guidance representations, RaysUp formulates an Any-Resolution Cross-Attention mechanism, enabling adaptive reconstruction of backbone features at arbitrary output resolutions.

To overcome the limitations imposed by fixed 2D local neighborhoods in JBU and to enhance the spatial kernel’s capacity to model underlying geometric structures, we further introduce Ray Positional Encoding (RayPE). Inspired by neural radiance field formulations \cite{eccv2020nerf}, RayPE encodes guidance features using 6D Plücker ray coordinates, thereby implicitly injecting 3D geometric priors into the attention module. This mechanism generalizes conventional joint bilateral upsampling into a geometry-aware reconstruction process, effectively alleviating boundary blurring and structural drift.

Finally, RaysUp adopts Geometry-Aware Neighborhood Attention as a learnable bilateral aggregation operator, performing localized cross-attention between RayPE-encoded high-resolution guidance features and low-resolution input features. This design preserves the locality property of JBU while enabling content-adaptive learning of upsampling weights, achieving a favorable balance among geometric fidelity, semantic consistency, and computational efficiency.

As illustrated in Fig. \ref{fig:teaser}, owing to our carefully designed architecture, RaysUp enables task-agnostic and VFM-agnostic feature upsampling with geometry-aware ray representations at arbitrary resolutions. Compared to the state-of-the-art method AnyUp, RaysUp achieves superior performance across multiple dense prediction tasks while using only approximately 16\% of the parameters. Moreover, it attains an approximately 7$\times$ improvement in inference efficiency, substantially reducing both computational and memory overhead, thereby demonstrating an advantageous trade-off between accuracy and efficiency.

In summary, our contributions are as follows:
\begin{itemize}
\item We propose RaysUp, an ultra-lightweight, task-agnostic, and VFM-agnostic framework for universal feature upsampling at arbitrary resolutions.

\item A lightweight Spatially Decoupled Guidance Encoder is proposed to extract direction-aware spatial semantic guidance features. Building upon this, RaysUp employs an Any-Resolution Cross-Attention mechanism to enable adaptive reconstruction of backbone features at arbitrary resolutions.

\item We propose Ray Positional Encoding (RayPE), which injects implicit 3D geometric priors via 6D Plücker ray coordinates, extending classical JBU into a geometry-aware reconstruction process that enhances boundary fidelity and mitigates structural drift.

\item Extensive experiments demonstrate that, with only about $1/5$ of the parameters of baseline methods, RaysUp achieves state-of-the-art performance across multiple dense prediction tasks and delivers approximately 7$\times$ faster inference, significantly reducing both computational and memory costs while balancing accuracy and practical efficiency.

\end{itemize}

\section{Related Work}
Feature upsampling aims to recover the spatial resolution of intermediate representations in deep neural networks and serves as a fundamental component in dense prediction tasks such as semantic segmentation \cite{jose2025dinov2txt, shin2024pixelclip, wysoczanska2024clipdinoiser, barsellotti2025talking} and depth estimation \cite{cvpr2024depthanything, yang2024da2, lin2026da3, chen2025videodepth}. Existing approaches can be broadly categorized into two paradigms: Discrete Reconstruction methods and Continuous Mapping methods.

\textbf{Discrete Reconstruction.}
This category of methods primarily relies on interpolation, filtering, or convolutional decoding mechanisms to explicitly reconstruct high-resolution features from low-resolution inputs. Classical interpolation strategies \cite{duchon1979lanczos, mckinley1998cubic}, such as nearest-neighbor and bilinear, are computationally efficient and flexible across scales; however, they lack content adaptivity and often result in boundary blurring and loss of fine details. To enhance structural preservation, Joint Bilateral Upsampling (JBU) \cite{KopfCLU07JBU} introduces a high-resolution guidance image and performs structure-aware reconstruction through the joint weighting of spatial and range kernels. Subsequently, adaptive filtering principles were incorporated into neural network frameworks. For instance, CARAFE \cite{Wang2019CARAFE} dynamically predicts content-aware reassembly kernels for feature reconstruction; SAPA \cite{Lu2022SAPA} improves spatial semantic consistency via similarity modeling; and FADE \cite{eccv2022fade} enhances representational capacity by introducing a semi-shift operator into the decoding structure. In addition, IndexNet \cite{pami2022indexnet} and A2U \cite{Dai2021a2u} have demonstrated strong performance in task-specific scenarios such as image matting. Despite these advances, such approaches typically depend on encoder–decoder architectures and downstream supervision, and are often tailored to specific backbones or fixed upsampling ratios, thereby limiting their generalization capability. Moreover, the classical JBU formulation remains constrained by fixed 2D neighborhoods and RGB-difference-based range kernels, which are insufficient to model complex semantic relationships and underlying geometric structures.

In contrast, RaysUp revisits the classical JBU framework and systematically reformulates its spatial and range kernels to overcome the limitations of fixed 2D neighborhoods and RGB-based range modeling. By introducing a spatially decoupled guidance encoder and an attention-based cross-scale interaction, RaysUp unifies filter-style local aggregation with adaptive feature interaction. This design enables resolution-flexible feature upsampling while improving structural fidelity and semantic expressiveness.

\textbf{Continuous Mapping.}
This line of research formulates feature upsampling as a resolution-agnostic continuous function approximation or cross-scale interaction problem, with an emphasis on enhancing representational capacity and scale flexibility. Inspired by implicit neural representations \cite{eccv2020nerf, Li_2025}, LiFT \cite{eccv2024lift} models image reconstruction as a coordinate-conditioned continuous function, enabling arbitrary-resolution prediction. Building upon this paradigm, FeatUp \cite{fu2024featup}  proposes a task-agnostic feature upsampling framework; however, its high-performance variant relies on per-image optimization, leading to complex training objectives and substantial deployment overhead. More recent approaches, such as JAFAR \cite{couairon2025jafar}  and LoftUp \cite{iccv2025loftup}, cast the upsampling process as cross-scale attention or coordinate-based mapping between high-resolution queries and low-resolution features, thereby achieving resolution-independent feature reconstruction. Although these methods improve expressiveness and scale generalization, they typically require retraining for specific visual encoders. Extending JAFAR, AnyUp \cite{wimmer2026anyup} demonstrates that by constraining the receptive field of attention, a universal upsampling paradigm can be constructed, allowing a single trained upsampler to generalize across multiple vision encoders with different feature dimensions without retraining.

Following this attention-based trajectory, RaysUp introduces a lightweight architecture containing only 0.14M parameters, which can be seamlessly integrated with arbitrary vision encoders and supports universal feature upsampling at arbitrary resolutions. Notably, RaysUp achieves approximately 7$\times$ faster inference than AnyUp, while substantially reducing memory consumption.

\begin{figure*}[!t]
    \centering
    \includegraphics[width=\linewidth]{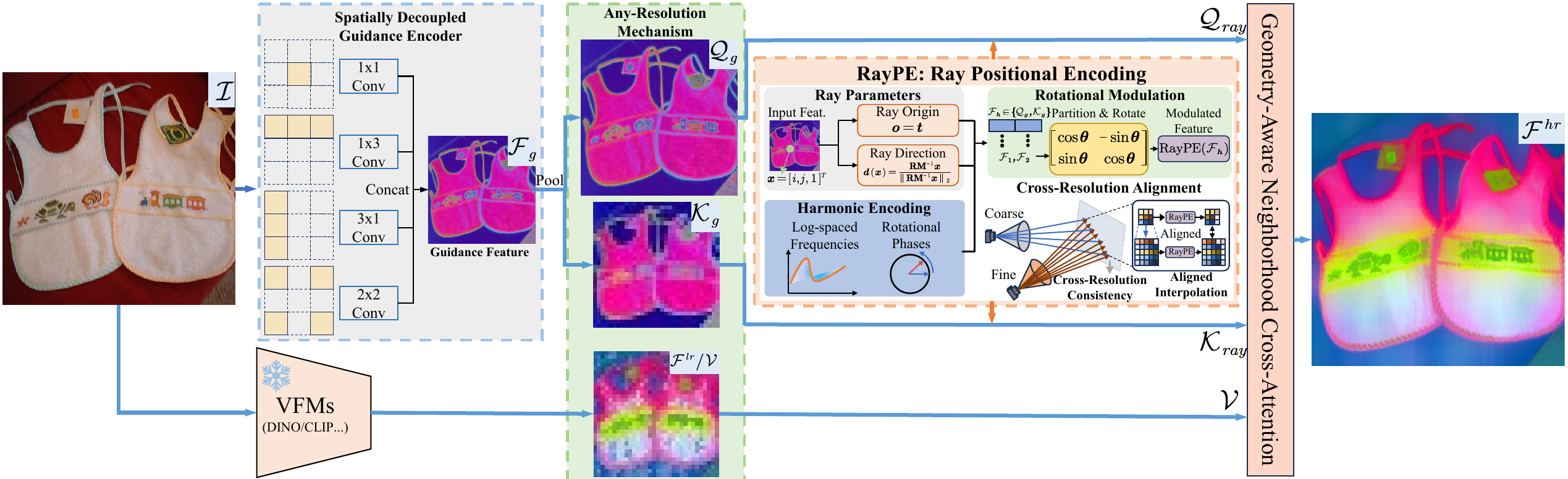}
    \caption{\textbf{Overview of RaysUp.} Given an image $\ten{I}$ and a low-resolution VFM feature map $\ten{F}^{lr}$, RaysUp reconstructs a high-resolution feature map $\ten{F}^{hr}$ at an arbitrary target resolution. A lightweight Spatially Decoupled Guidance Encoder first extracts direction-aware guidance features $\ten{F}_g$, which are adaptively pooled to generate target-resolution queries $\ten{Q}_g$ and VFM-resolution keys $\ten{K}_g$. After RayPE, the resulting geometry-aware ray representations $\ten{Q}_{ray}$ and $\ten{K}_{ray}$ inject implicit 3D spatial priors into the attention computation. Finally, a geometry-aware Neighborhood Cross-Attention module aggregates low-resolution features $\ten{F}^{lr}$ as values to produce the reconstructed high-resolution features, ensuring cross-resolution spatial consistency while preserving underlying 3D geometric structure.}
    \label{fig:framework}
    \vspace{-10px}
\end{figure*}
\section{RaysUp}
\subsection{Overview}
As illustrated in Fig. \ref{fig:framework}, given an input RGB image $\ten{I} \in \mathbb{R}^{3 \times H_{in} \times W_{in}}$ and a low-resolution feature map $\ten{F}^{lr} \in \mathbb{R}^{D_f \times H_{lr} \times W_{lr}}$ extracted from a Vision Foundation Model (VFM), where $D_f$ denotes the feature dimension and $(H_{lr}, W_{lr})$ are typically small (e.g., $16 \times 16$ or $32 \times 32$), the goal of RaysUp is to reconstruct a high-resolution feature map $\ten{F}^{hr} \in \mathbb{R}^{D_f \times H_{any} \times W_{any}}$ at an arbitrary target resolution $(H_{any}, W_{any})$. To achieve this, RaysUp first employs a lightweight Spatially Decoupled Guidance Encoder to extract direction-aware guidance features from the input image, producing $\ten{F}_g \in \mathbb{R}^{D_g \times H_{in} \times W_{in}}$, where $D_g$ denotes the dimensionality of the guidance features. Then, an adaptive average pooling operation generates a target-resolution query 
$\ten{Q}_g \in \mathbb{R}^{D_g \times H_{any} \times W_{any}}$ and a VFM-resolution key $\ten{K}_g \in \mathbb{R}^{D_g \times H_{lr} \times W_{lr}}$. Both $\ten{Q}_g$ and $\ten{K}_g$ are subsequently encoded via Ray Positional Encoding (RayPE) to produce geometry-aware ray representations $\ten{Q}_{ray}$ and $\ten{K}_{ray}$, which embed implicit 3D spatial priors. Finally, a geometry-aware Neighborhood Cross-Attention mechanism aggregates the low-resolution features $\ten{F}^{lr}$ (acting as the value $\ten{V}$) into the high-resolution positions, yielding the reconstructed feature map $\ten{F}^{hr}$. This process ensures spatial consistency across resolutions while preserving the underlying 3D geometric structure.

\subsection{Lightweight Spatially Decoupled Guidance Encoder}
Guidance encoders for dense feature extraction conventionally adopt standard $1\times1$ or $3\times3$ convolutional operators \cite{couairon2025jafar, wimmer2026anyup, iccv2025loftup}. However, statistical analysis of the learned $3\times3$ convolutional kernels in JAFAR \cite{couairon2025jafar} reveals a markedly anisotropic spatial distribution (Appendix \ref{sec:motivation_sdge}). Specifically, the average kernel magnitude matrix exhibits consistently larger weights at the four corner positions, while the central cross region (i.e., horizontal and vertical axes) shows relatively smaller magnitudes. This observation indicates that local feature aggregation is not spatially isotropic.

Motivated by this empirical regularity, we design a lightweight spatially decoupled guidance encoder to enhance directional interpretability and structural disentanglement. Concretely, we decompose the implicit receptive field into multiple orthogonal directional components and model them in parallel. The total output channel dimension $D_g$ is evenly divided across branches, such that each branch is responsible for learning only $D_g/4$ channels. Given an input image $\ten{I} \in \mathbb{R}^{3 \times H_{in} \times W_{in}}$, the branch-wise formulations are defined as:
\begin{equation}
\begin{aligned}
    \ten{F}^c_g &= \phi \left(\mathrm{GN} \left(\mathrm{Conv}_{1\times1}^{D_g/4}(\ten{I})\right)\right),
    \ten{F}^h_g = \phi \left(\mathrm{GN} \left(\mathrm{Conv}_{1\times3}^{D_g/4}(\ten{I})\right)\right),\\
    \ten{F}^v_g &= \phi \left(\mathrm{GN} \left(\mathrm{Conv}_{3\times1}^{D_g/4}(\ten{I})\right)\right),
    \ten{F}^{diag}_g = \phi \left(\mathrm{GN} \left(\mathrm{Conv}_{2\times2,d=2}^{D_g/4}(\ten{I})\right)\right),
\end{aligned}
\end{equation}
where $\mathrm{Conv}_{k\times k}^{D_g/4}$ denotes a convolution with $D_g/4$ output channels, and $\mathrm{Conv}_{2\times2,d=2}$ represents a $2\times2$ dilated convolution with dilation rate 2. $\mathrm{GN}(\cdot)$ denotes Group Normalization \cite{wu2018groupnorm}, and $\phi(\cdot)$ is the SiLU activation function \cite{elfwing2018silu}. To further enhance representational capacity, each branch adopts a shallow residual structure:
\begin{equation}
    \tilde{\ten{F}}^b_g = \ten{X}^b + \phi \left(\mathrm{GN} \left(\mathrm{Conv}_b(\ten{X}^b)\right)\right),
\quad b \in \{c,h,v,diag\},
\end{equation}
where $\ten{X}^b$ denotes the output of the first convolution in branch $b$. Finally, the directional features are concatenated along the channel dimension to form the guidance feature representation:
\begin{equation}
    \ten{F}_g(\ten{I}) = \mathrm{Concat}(\tilde{\ten{F}}^c_g, \tilde{\ten{F}}^h_g, \tilde{\ten{F}}^v_g, \tilde{\ten{F}}^{diag}_g)
\in \mathbb{R}^{D_g\times H_{in} \times W_{in}}.
\end{equation}

From a parameterization standpoint, a standard $3\times3$ convolution contains $27D_g$ parameters, whereas our proposed spatially decoupled encoder requires only $3 \cdot \frac{D_g}{4} (1 + 3 + 3 + 4)= 8.25D_g$. Accordingly, the spatially decoupled guidance encoder not only substantially enhances anisotropic modeling capability but also reduces parameter overhead by approximately 69.4\%, demonstrating clear advantages in both parameter efficiency and computational cost.

\subsection{Any-Resolution Cross-Attention Mechanism}
To enable feature upsampling at arbitrary resolutions, we adopt a resolution-decoupled cross-attention mechanism \cite{couairon2025jafar}. Specifically, the guidance encoded features $\ten{F}_g(\ten{I})$ are used to construct multi-scale attention inputs. Queries $\ten{Q}_g \in \mathbb{R}^{D_g \times H_{any} \times W_{any}}$ are generated at the upsampled resolution:
\begin{equation}
    \ten{Q}_g = \text{AdaptiveAvgPool}(\ten{F}_g(\ten{I}), H_{any}, W_{any}),
\end{equation}
while keys $\ten{K}_g \in \mathbb{R}^{D_g \times H_{lr} \times W_{lr}}$ are produced at the backbone feature resolution:
\begin{equation}
    \ten{K}_g = \text{AdaptiveAvgPool}(\ten{F}_g(\ten{I}), H_{lr}, W_{lr}),
\end{equation}
and the value features are taken directly from the output of VFMs:
\begin{equation}
    \ten{V} = \ten{F}^{lr}  \in \mathbb{R}^{D_f \times H_{lr} \times W_{lr}}.
\end{equation}

This mechanism enables high-resolution queries to aggregate information from semantically aligned low-resolution features without requiring explicit interpolation, thereby supporting flexible cross-resolution feature propagation.

\subsection{RayPE: Ray Positional Encoding}
Traditional feature upsampling methods \cite{wimmer2026anyup, couairon2025jafar, iccv2025loftup, fu2024featup, eccv2024lift} operate within the 2D image grid, implicitly assuming that Euclidean proximity reflects 3D geometric consistency. However, under perspective projection, adjacent pixels in the image may correspond to spatially distant 3D points (e.g., at depth discontinuities), whereas distant pixels may lie on the same physical surface. Therefore, spatial interpolation in image space cannot guarantee geometric consistency.

Inspired by Neural Radiance Fields (NeRF) \cite{eccv2020nerf}, RaysUp lifts feature reconstruction into a projective ray domain by associating each pixel with its camera ray. Let each pixel correspond to a ray $\vect{r}(\vect{x}) = (\vect{o}, \vect{d}(\vect{x}))$, where $\vect{o}$ is the camera origin and $\vect{d}(\vect{x})$ is the normalized direction. In this formulation, the upsampling can then be interpreted as transporting features from a coarse sampling of rays (backbone resolution) to a denser sampling (target resolution).

However, directly adopting a NeRF-style rendering pipeline incurs substantial computational cost. To address this, RaysUp introduces Ray Positional Encoding (RayPE), which encodes guidance features by mapping camera geometry into a high-dimensional rotational space. This modulation aligns features along the 3D ray manifold during upsampling. Unlike conventional 2D positional encodings \cite{su2024rope, icml2024sinrays}, RayPE explicitly binds feature aggregation to the underlying 3D ray geometry rather than relying on pixel-plane proximity.

\textbf{Ray Geometry Parameterization.}
Given a pixel $\vect{x}=[i,j,1]^T$ in guidance features 
$\ten{Q}_g \in \mathbb{R}^{D_g \times H_{any} \times W_{any}}$ 
or 
$\ten{K}_g \in \mathbb{R}^{D_g \times H_{lr} \times W_{lr}}$, 
the corresponding 3D ray is uniquely determined by the camera intrinsic matrix $\mat{M}$ and extrinsic parameters $\mat{T}=[\mat{R}|\vect{t}]$. The ray origin (camera center) and normalized direction in world coordinates are defined as:
\begin{equation}
    \vect{o} = \vect{t}, 
    \quad
    \vect{d}(\vect{x}) = 
    \frac{\mat{R} \mat{M}^{-1} \vect{x}}
    {\|\mat{R} \mat{M}^{-1} \vect{x}\|_2}.
\end{equation}
Then, the origin and direction are concatenated to form a unified ray descriptor:
\begin{equation}
    \vect{r} = \text{Concat}(\vect{o}, \vect{d}(\vect{x})) 
    \in \mathbb{R}^{6}.
\end{equation}

This parameterization provides a resolution-independent 3D geometric reference shared across feature maps.

\textbf{Multi-band Harmonic Phase Encoding.}
To capture geometric variations at multiple spatial scales, 
we construct a log-spaced frequency set 
$\{\omega_f\}_{f=0}^{N-1}$:
\begin{equation}
    \omega_f = 
    \exp\Bigg(
    \log\frac{2\pi}{\lambda_{\max}} 
    + \frac{f}{N-1}
    \Big(
    \log\frac{2\pi}{\lambda_{\min}} 
    - \log\frac{2\pi}{\lambda_{\max}}
    \Big)
    \Bigg),
\end{equation}
where $N$ denotes the total number of frequency bands, index $f \in \{0,\dots,N-1\}$ specifies the $f$-th frequency level, and $\lambda_{\max}$ and $\lambda_{\min}$ represent the maximum and minimum spatial wavelengths, which determine the lowest and highest angular frequencies in the encoding. For each spatial location, the rotational phase $\mathbf{\Theta} \in \mathbb{R}^{N \times 6}$ is computed by multiplying each frequency $\omega_f$ with the corresponding component $r_d$ of ray  descriptor $\vect{r}$:
\begin{equation}
    \mathbf{\Theta}_{f,d} = \omega_f \, r_d,
    \quad
    f=0,\dots,N-1,\;
    d=1,\dots,6.
\end{equation}
Then, the frequency and coordinate axes are flattened into a phase vector:
\begin{equation}
    \boldsymbol{\theta} = \mathrm{Flatten}(\mathbf{\Theta}) 
    \in \mathbb{R}^{6N},
\end{equation}
which encodes multi-scale geometric information along the ray, providing a unified reference for aligning high- and low-resolution features in 3D space.

\textbf{Ray-driven Rotational Modulation.}
For each attention-head guidance feature tensor 
$\ten{F}_h \in \{\ten{Q}_g, \ten{K}_g\}$ 
with dimensionality $d_{\text{head}}$, we first split the feature into two halves:
\begin{equation}
    \ten{F}_h = [\ten{F}_1, \ten{F}_2], 
    \quad 
    \ten{F}_1, \ten{F}_2 \in \mathbb{R}^{\lfloor d_{\text{head}}/2 \rfloor}.
\end{equation}
To accommodate the rotate-half structure of attention heads, we require $6N \le \lfloor d_{\text{head}}/2\rfloor$, where $6N$ corresponds to the number of flattened RayPE phases. If $6N < \lfloor d_{\text{head}}/2\rfloor$, the remaining feature dimensions are padded with zero phase, equivalent to an identity rotation.
Following the idea of RoPE \cite{su2024rope}, Ray-driven rotational modulation is applied element-wise to each pair of feature dimensions:
\begin{equation}
\begin{bmatrix}
\tilde{\ten{F}}_1 \\
\tilde{\ten{F}}_2
\end{bmatrix}
=
\begin{bmatrix}
\cos\boldsymbol{\theta} & -\sin\boldsymbol{\theta} \\
\sin\boldsymbol{\theta} & \cos\boldsymbol{\theta}
\end{bmatrix}
\begin{bmatrix}
\ten{F}_1 \\
\ten{F}_2
\end{bmatrix}.
\end{equation}
Finally, the RayPE encoded feature is obtained by concatenation:
\begin{equation}
    \tilde{\ten{F}}_h = \text{Concat}(\tilde{\ten{F}}_1, \tilde{\ten{F}}_2).
\end{equation}
As the rotation angles are computed from the 3D ray coordinates and multi-scale harmonic frequencies, RayPE extends RoPE \cite{su2024rope} from 1D sequential positions to 3D geometric positions, ensuring alignment of cross-resolution features along the shared 3D ray manifold and enforcing geometric consistency during upsampling.

\subsection{Geometry-Aware Neighborhood Cross-Attention}
To reconstruct dense features while preserving geometric consistency, we employ geometry-aware Neighborhood Cross-Attention \cite{hassani2023nca}. Instead of global attention, each high-resolution query aggregates information from a local neighborhood defined in the low-resolution feature space. Let $\ten{Q}_{ray}$ and $\ten{K}_{ray}$ denote RayPE-modulated queries and keys, respectively, and $\ten{V}$ the value features (VFM features). For a query position $(i,j)$, feature reconstruction is formulated as:
\begin{equation}
    \ten{F}^{hr}_{i,j}=
\sum_{(u,v)\in\mathcal{N}_{i,j}}
\text{Softmax}
\left(
\frac{(\ten{Q}_{ray})_{i,j}^{\top}(\ten{K}_{ray})_{u,v}}{\sqrt{d}}
\right)
\ten{V}_{u,v},
\end{equation}
where $\mathcal{N}_{i,j}$ denotes a $k \times k$ local neighborhood centered at the corresponding key location.
To bridge resolution discrepancies between high-resolution $(H_{any},W_{any})$ queries and low-resolution $(H_{lr},W_{lr})$ keys, dilation factors are defined as:
\begin{equation}
    d_h = \max(1,\lfloor H_{any}/H_{lr} \rfloor),
d_w = \max(1,\lfloor W_{any}/W_{lr} \rfloor).
\end{equation}
The dilated neighborhood induces a locally consistent sampling on the underlying ray manifold, ensuring that attention is performed among rays with similar orientations. Consequently, feature aggregation respects angular consistency, preventing interactions between geometrically unrelated rays while maintaining cross-resolution correspondence.

Under a local planar approximation, this process can be interpreted as discrete feature propagation along smoothly varying ray directions, enabling geometry-consistent cross-resolution information transfer. Meanwhile, the computational complexity is reduced from global attention $\mathcal{O}(H_{any}W_{any} \cdot H_{lr}W_{lr})$ to $\mathcal{O}(H_{any}W_{any} \cdot k^2)$, thereby improving efficiency while preserving geometric consistency.

\subsection{Training Objective}
Following standard practices for feature upsampling \cite{couairon2025jafar, wimmer2026anyup}, RaysUp is trained with a reconstruction objective. Given a high-resolution image $\ten{I}_{hr}$, a low-resolution counterpart $\ten{I}_{lr}$ is generated by downsampling $\ten{I}_{hr}$ with a random scale factor $s \in [2,4]$. Both images are then fed into a frozen vision encoder to obtain the target high-resolution features $\ten{F}_{tgt} \in \mathbb{R}^{D_f \times H_{any} \times W_{any}}$ and the source low-resolution features $\ten{F}^{lr} \in \mathbb{R}^{D_f \times H_{lr} \times W_{lr}}$. Conditioned on $\ten{F}^{lr}$ and the high-resolution guidance image $\ten{I}_{hr}$, the RaysUp upsampler predicts the reconstructed high-resolution feature map $\hat{\ten{F}}^{hr} \in \mathbb{R}^{D_f \times H_{any} \times W_{any}}$.

To enforce feature consistency, the training loss $\mathcal{L}$ is defined as a combination of cosine similarity loss $\mathcal{L}_{cos}$ and $L_2$ distance loss $\mathcal{L}_{L2}$:
\begin{equation}
    \mathcal{L} = \mathcal{L}_{cos}(\hat{\ten{F}}^{hr}, \ten{F}_{tgt}) + \mathcal{L}_{L2}(\hat{\ten{F}}^{hr}, \ten{F}_{tgt}).
\end{equation}

\section{Experiments}
\subsection{Experimental Setup}
\textbf{Implementation.}
RaysUp was trained on the standard ImageNet dataset \cite{cvpr2009imagenet}. The model was optimized using the AdamW optimizer for 100,000 iterations with a batch size of 4 and an initial learning rate of $2 \times 10^{-4}$. To construct training pairs, target high-resolution images were uniformly resized to a spatial resolution of $448 \times 448$. For improved training efficiency, the guidance input to the upsampler was downsampled to $224 \times 224$. Owing to the adopted training strategy and the lightweight architectural design, the entire training process completed in approximately one hour on a single NVIDIA A100 GPU. Additional implementation details were provided in the supplementary material.

\textbf{Baselines.}
To comprehensively evaluate the performance and generalization capability of RaysUp, it was systematically compared against several representative methods, including: traditional interpolation (e.g., Bilinear), fixed-ratio task-agnostic methods requiring encoder-specific retraining (e.g., FeatUp \cite{fu2024featup}, LoftUp \cite{iccv2025loftup}), arbitrary-resolution encoder-dependent methods (e.g., JAFAR \cite{couairon2025jafar}), and arbitrary-resolution encoder-agnostic methods (e.g., AnyUp \cite{wimmer2026anyup}).

\subsection{Task-agnostic Performance}
Following the baseline protocol \cite{fu2024featup,couairon2025jafar, wimmer2026anyup, iccv2025loftup}, RaysUp employed DINOv2-S \cite{Oquab24dinov2} as the default backbone for both training the upsampler and extracting features for downstream tasks.

\textbf{Semantic Segmentation.}
Following the experimental protocol of JAFAR \cite{couairon2025jafar}, RaysUp employed a $1\times1$ convolutional layer as a linear probe for semantic label prediction. Evaluation was conducted on COCO-Stuff \cite{caesar2018coco}, ADE20K \cite{zhou2019ade20k}, Pascal-VOC \cite{ijcv2015voc}, and Cityscapes \cite{Cordts2016cityscapes} datasets. Performance was reported in terms of mean Intersection over Union (mIoU) and pixel accuracy (Acc).

\textbf{Depth and Surface Normal Estimation.}
Following Probe3D \cite{el2024probing3d}, RaysUp was evaluated on dense geometric prediction tasks using NYUv2 \cite{eccv2012nyuv2}. For both normal and depth estimation, RMSE served as the primary metric. For normals, we additionally reported accuracy under angular thresholds of $11.25^\circ$, $22.5^\circ$, and $30^\circ$. For depth, we reported the $\delta_1$ metric, which measured the proportion of pixels with a prediction-to-ground-truth ratio below $1.25$.

\textbf{Video Object Segmentation and Open-Vocabulary Segmentation.}
RaysUp was evaluated on the DAVIS \cite{pont2017davis} dataset for video object segmentation. Specifically, the first-frame object mask was propagated temporally by computing dense cross-frame feature similarities. Performance was measured using $\mathcal{J}$ Mean (region similarity IoU), $\mathcal{F}$ Mean (boundary contour accuracy), and their combined metric, $\mathcal{J}\&\mathcal{F}$ Mean, providing a comprehensive assessment of regional consistency and boundary reconstruction quality. For zero-shot open-vocabulary segmentation, the upsampled features were integrated into the ProxyCLIP \cite{lan2024proxyclip} framework. By leveraging spatial correspondences from pre-trained vision encoders to enhance CLIP \cite{Radford21CLIP} representations, we evaluated segmentation performance on COCO-Stuff \cite{caesar2018coco}, Cityscapes \cite{Cordts2016cityscapes}, ADE20K \cite{zhou2019ade20k}, and Pascal-VOC \cite{ijcv2015voc}.

\begin{table*}[t!]
\centering
\caption{\textbf{Task-agnostic Performance.}}
\label{tab:all_tasks}
\resizebox{\linewidth}{!}{
\begin{tabular}{l c c c c c c c c c c c}
    \toprule

    \multicolumn{12}{c}{\textbf{Semantic Segmentation}} \\
    \specialrule{1.2pt}{2pt}{2pt}

    & \multirow{2}{*}{Method} & \multicolumn{2}{c}{COCO-Stuff \cite{caesar2018coco}} 
    & \multicolumn{2}{c}{Pascal-VOC \cite{ijcv2015voc}} 
    & \multicolumn{2}{c}{ADE20K \cite{zhou2019ade20k}} 
    & \multicolumn{2}{c}{Cityscape \cite{Cordts2016cityscapes}}
    & \multicolumn{2}{c}{} \\
    \cmidrule(lr){3-4}
    \cmidrule(lr){5-6}
    \cmidrule(lr){7-8}
    \cmidrule(lr){9-10}

    &   
    & mIoU$\uparrow$ & Acc$\uparrow$ 
    & mIoU$\uparrow$ & Acc$\uparrow$ 
    & mIoU$\uparrow$ & Acc$\uparrow$ 
    & mIoU$\uparrow$ & Acc$\uparrow$ 
    & & \\

    \cmidrule(lr){2-10}
    & Bilinear 
    & 59.58 & 79.42 
    & 81.70 & 95.44 
    & 40.47 & 74.13 
    & 59.72 & 92.55
    & & \\

    & FeatUp \cite{fu2024featup}
    & 61.89 & 81.10 
    & 83.37 & 96.01 
    & \underline{42.33} & 75.65 
    & 60.18 & 93.05
    & & \\

    & LoftUp \cite{iccv2025loftup}
    & \underline{62.23} & \underline{81.38} 
    & \underline{84.50} & \underline{96.30} 
    & 42.17 & \underline{75.79} 
    & \textbf{62.09} & \underline{93.54}
    & & \\

    & JAFAR \cite{couairon2025jafar}
    & 61.79 & 81.11 
    & 83.89 & 96.11 
    & 42.16 & 75.56 
    & 61.40 & 93.47
    & & \\

    & AnyUp \cite{wimmer2026anyup}
    & 62.14 & \underline{81.38} 
    & 84.18 & 96.20 
    & 42.15 & 75.71 
    & 60.62 & 93.26
    & & \\

    & \textbf{RaysUp}
    & \textbf{62.32} & \textbf{81.47} 
    & \textbf{84.64} & \textbf{96.34} 
    & \textbf{42.34} & \textbf{75.81} 
    & \underline{61.88} & \textbf{93.64}
    & & \\

    \specialrule{1.2pt}{3pt}{3pt}
    \multicolumn{12}{c}{\textbf{Surface Normal and Depth Estimation}} \\
    \specialrule{1.2pt}{2pt}{2pt}

    & \multirow{2}{*}{Method} & \multicolumn{4}{c}{Surface Normal} 
    & \multicolumn{2}{c}{Depth (Abs)} 
    & \multicolumn{2}{c}{Depth (Rel)}
    & \multicolumn{2}{c}{} \\
    \cmidrule(lr){3-6}
    \cmidrule(lr){7-8}
    \cmidrule(lr){9-10}

    & 
    & RMSE$\downarrow$ & $11.25^\circ\uparrow$ & $22.5^\circ\uparrow$ & $30^\circ\uparrow$
    & RMSE$\downarrow$ & $\delta_1\uparrow$
    & RMSE$\downarrow$ & $\delta_1\uparrow$
    & & \\

    \cmidrule(lr){2-10}
    & Bilinear
    & 28.23 & 0.4933 & 0.6974 & 0.7718
    & 0.4789 & 0.7987
    & 0.3348 & 0.9243
    & & \\

    & FeatUp \cite{fu2024featup}
    & 28.94 & 0.4846 & 0.6888 & 0.7619
    & 0.4781 & \textbf{0.8117}
    & 0.3393 & 0.9205
    & & \\

    & LoftUp \cite{iccv2025loftup}
    & 28.45 & 0.4861 & 0.6931 & 0.7689
    & 0.4828 & 0.7958
    & 0.3353 & 0.9221
    & & \\

    & JAFAR \cite{couairon2025jafar}
    & \underline{27.80} & 0.4948 & 0.6986 & 0.7740
    & \underline{0.4693} & 0.8071
    & 0.3255 & 0.9301
    & & \\

    & AnyUp \cite{wimmer2026anyup}
    & 27.83 & \underline{0.4962} & \underline{0.7014} & \underline{0.7767}
    & 0.4781 & 0.8019
    & \underline{0.3244} & \underline{0.9306}
    & & \\

    & \textbf{RaysUp}
    & \textbf{27.69} & \textbf{0.4986} & \textbf{0.7030} & \textbf{0.7775}
    & \textbf{0.4658} & \underline{0.8103}
    & \textbf{0.3195} & \textbf{0.9309}
    & & \\

    \specialrule{1.2pt}{3pt}{3pt}
    \multicolumn{12}{c}{\textbf{Video Object Segmentation and Open-Vocabulary Segmentation}} \\
    \specialrule{1.2pt}{2pt}{2pt}

    & \multicolumn{3}{c}{Video Obj. Seg.} 
    & \multicolumn{8}{c}{Open-Voc Seg.} \\

    \cmidrule(lr){2-4}
    \cmidrule(lr){5-12}

    Method 
    & \multirow{2}{*}{$\mathcal{J}\uparrow$} & \multirow{2}{*}{$\mathcal{F}\uparrow$} & \multirow{2}{*}{$\mathcal{J\&F}\uparrow$ }
    & \multicolumn{2}{c}{COCO-Stuff \cite{caesar2018coco}}
    & \multicolumn{2}{c}{Pascal-VOC \cite{ijcv2015voc}}
    & \multicolumn{2}{c}{ADE20K \cite{zhou2019ade20k}}
    & \multicolumn{2}{c}{Cityscape \cite{Cordts2016cityscapes}} \\

    \cmidrule(lr){5-6}
    \cmidrule(lr){7-8}
    \cmidrule(lr){9-10}
    \cmidrule(lr){11-12}

    & & & & mIoU$\uparrow$ & Acc$\uparrow$ 
    & mIoU$\uparrow$ & Acc$\uparrow$ 
    & mIoU$\uparrow$ & Acc$\uparrow$ 
    & mIoU$\uparrow$ & Acc$\uparrow$  \\

    \midrule

    Bilinear 
    & 63.21 & 66.53 & 64.87 
    & 25.73 & 47.91 
    & 59.71 & 85.87 
    & 19.60 & 42.32 
    & 34.91 & 58.52 \\

    FeatUp \cite{fu2024featup}   
    & 65.45 & 72.03 & 68.74 
    & 26.67 & 48.13 
    & 54.96 & 84.02 
    & 19.78 & \underline{43.09} 
    & 35.23 & 57.44 \\

    LoftUp \cite{iccv2025loftup}  
    & 67.32 & 74.53 & 70.92 
    & \textbf{27.54} & 48.38 
    & \textbf{65.21} & \textbf{87.89} 
    & \textbf{21.10} & \textbf{43.28} 
    & \textbf{38.25} & 59.95 \\

    JAFAR \cite{couairon2025jafar}   
    & 67.18 & \underline{74.62} & 70.90 
    & 27.13 & 48.47 
    & 63.52 & 87.43 
    & \underline{20.57} & 42.76 
    & \underline{38.00} & \textbf{60.32} \\

    AnyUp \cite{wimmer2026anyup}   
    & \underline{67.55} & 74.42 & \underline{70.98} 
    & \underline{27.30} & \textbf{48.62} 
    & \underline{63.67} & \underline{87.50} 
    & 20.67 & 42.77 
    & 37.56 & \underline{60.03} \\

    \textbf{RaysUp} 
    & \textbf{68.14} & \textbf{74.81} & \textbf{71.47} 
    & 27.11 & \underline{48.57} 
    & 63.28 & 87.34 
    & 20.54 & 42.71 
    & 37.24 & 59.68 \\
    \bottomrule
\end{tabular}}
\vspace{-10px}
\end{table*}

\begin{figure*}
    \centering
    \includegraphics[width=\linewidth]{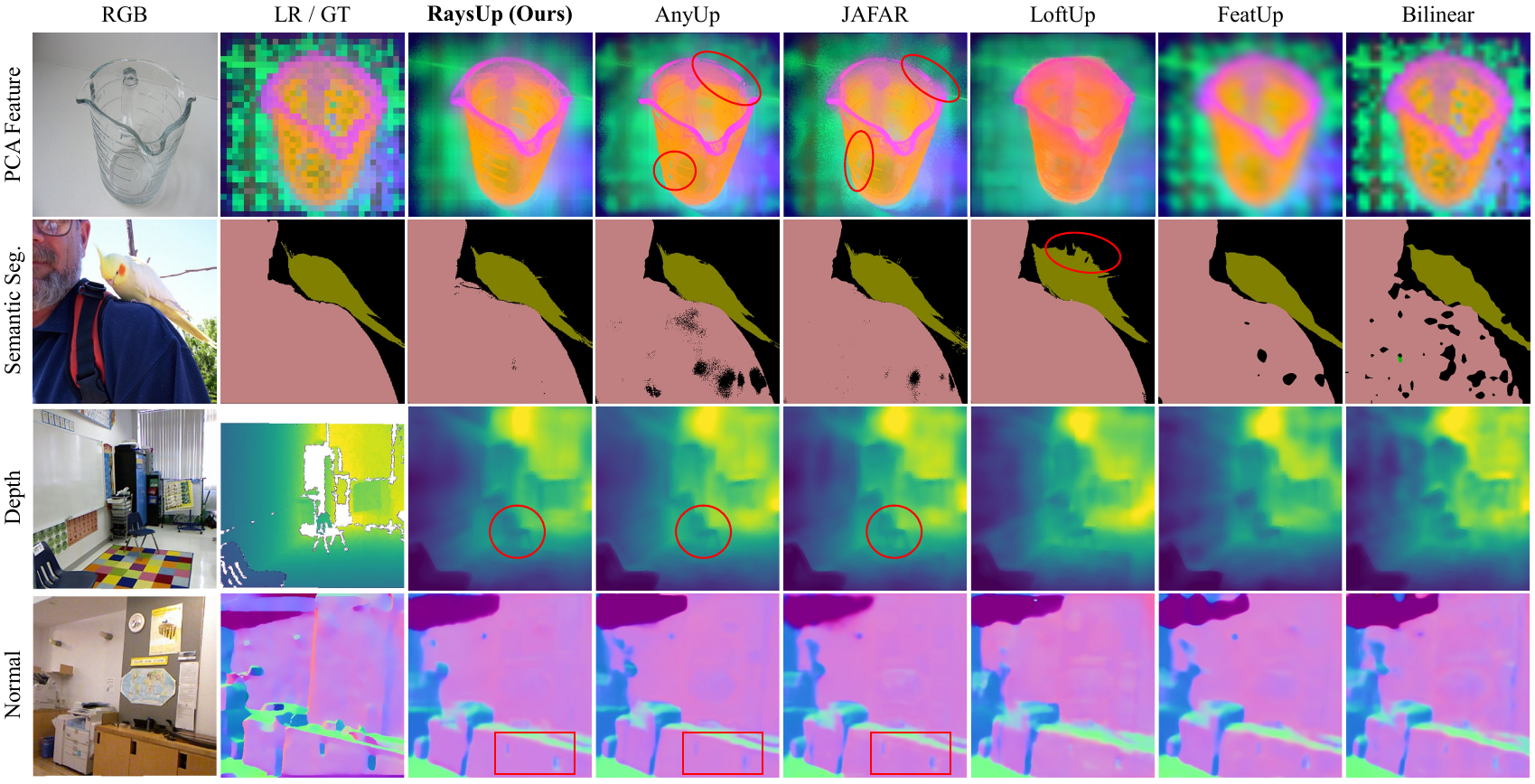}
   \caption{\textbf{Qualitative Results for Task-agnostic Performance.} Leveraging the geometry-aware Ray representation, RaysUp demonstrates enhanced geometric consistency and achieves superior performance across downstream tasks.}
    \label{fig:task_baseline}
\end{figure*}

\textbf{Analysis.}
The quantitative results (Tab. \ref{tab:all_tasks}) indicate that, owing to our Geometry-Aware Ray Representation, RaysUp achieves the best or second-best performance across most metrics, with particularly strong results in dense geometric prediction tasks such as surface normal and depth estimation. In comparison, LoftUp \cite{iccv2025loftup} shows slightly superior performance in certain semantic segmentation and open-vocabulary tasks, primarily due to the incorporation of SAM \cite{kirillov2023sam} masks as auxiliary supervision during training. Qualitative analysis (Fig. \ref{fig:task_baseline}) further demonstrates that, unlike AnyUp \cite{wimmer2026anyup}, JAFAR \cite{couairon2025jafar}, and FeatUp \cite{fu2024featup} which exhibit holes, or LoftUp \cite{iccv2025loftup} which produces blurred boundaries, RaysUp better preserves geometric consistency. Overall, even without additional segmentation supervision, RaysUp maintains overall superior performance, with pronounced advantages in geometric consistency and video tasks, highlighting its stronger task-agnostic generalization capability.

\begin{table}[!ht]
    \centering
   \caption{\textbf{VFM-agnostic Performance.} Semantic segmentation results (mIoU/Acc) on Pascal-VOC \cite{ijcv2015voc}  and depth estimation results (RMSE) on NYUv2 \cite{eccv2012nyuv2}.}
    \label{tab:encoder_generalization_scales}
    \resizebox{0.9\textwidth}{!}{
    \begin{tabular}{l l | c c | c c}
        \toprule
        & & \multicolumn{2}{c|}{DINOv2} & \multicolumn{2}{c}{DINOv3} \\
        \cmidrule(lr){3-4} \cmidrule(l){5-6}
        Size & Method & mIoU (Acc) $\uparrow$ & RMSE (abs/rel) $\downarrow$ & mIoU (Acc) $\uparrow$ & RMSE (abs/rel) $\downarrow$ \\
        \midrule
        \multirow{2}{*}{ViT-S} & AnyUp \cite{wimmer2026anyup}  & 84.18 (96.20) & 0.478 (0.324) & 81.97 (95.77) & 0.506 (0.345) \\
              & \textbf{RaysUp} & \textbf{84.64 (96.34)} & \textbf{0.465 (0.319)} & \textbf{82.60 (95.97)} & \textbf{0.502 (0.342)} \\
        \cmidrule{1-6}
        \multirow{2}{*}{ViT-M} & AnyUp \cite{wimmer2026anyup} & 85.44 (96.57) & 0.415 (0.289) & 86.51 (96.84) & 0.455 (0.296) \\
              & \textbf{RaysUp} & \textbf{86.03 (96.70)} & \textbf{0.400 (0.276)} & \textbf{87.20 (97.02)} & \textbf{0.452 (0.289)} \\
        \cmidrule{1-6}
        \multirow{2}{*}{ViT-L} & AnyUp  \cite{wimmer2026anyup} & 85.47 (96.56) & 0.393 (0.259) & 87.48 (97.07) & 0.402 (0.257) \\
              & \textbf{RaysUp} & \textbf{86.33 (96.81)} & \textbf{0.376 (0.256)} & \textbf{88.07 (97.20)} & \textbf{0.398 (0.255)} \\
        
        \midrule\midrule
        
        & & \multicolumn{2}{c|}{SigLIP2$^\dagger$} & \multicolumn{2}{c}{PE Spatial} \\
        \cmidrule(lr){3-4} \cmidrule(l){5-6}
        Size & Method & mIoU (Acc) $\uparrow$ & RMSE (abs/rel) $\downarrow$ & mIoU (Acc) $\uparrow$ & RMSE (abs/rel) $\downarrow$ \\
        \midrule
        \multirow{2}{*}{ViT-B} & AnyUp \cite{wimmer2026anyup} & 78.70 (94.60) & 0.845 (0.522) & 77.41 (94.63) & 0.726 (0.461) \\
              & \textbf{RaysUp} & \textbf{79.37 (94.84)} & \textbf{0.843 (0.519)} & \textbf{77.47 (94.69)} & \textbf{0.706 (0.448)} \\
        \cmidrule{1-6}
        \multirow{2}{*}{ViT-L} & AnyUp \cite{wimmer2026anyup} & 79.16 (94.73) & 0.726 (0.474) & 82.23 (95.87) & 0.635 (0.404) \\
              & \textbf{RaysUp} & \textbf{80.10 (94.97)} & \textbf{0.725 (0.472)} & \textbf{82.75 (95.98)} & \textbf{0.613 (0.388)} \\
        \bottomrule
    \end{tabular}}
\parbox{0.9\textwidth}{\footnotesize
$^\dagger$ A ViT-S version of SigLIP2 is unavailable.}
    \vspace{-10px}
\end{table}

\subsection{VFM-agnostic Performance}
Among the baseline methods, only AnyUp \cite{wimmer2026anyup} possesses VFM-agnostic capability. Therefore, it was selected as the primary comparison. Consistent with AnyUp, RaysUp was trained on DINOv2-S and evaluated for its generalization across different VFMs (DINOv2 \cite{Oquab24dinov2}, DINOv3 \cite{simeoni2025dinov3}, SigLIP2 \cite{tschannen2025siglip2}, and PE Spatial \cite{bolya2025pespatial}) and various model scales (ViT-S, ViT-M, and ViT-L). As shown in \cref{tab:encoder_generalization_scales}, RaysUp consistently outperforms AnyUp across all tested architectures and model sizes, demonstrating that the proposed ray-based upsampling approach effectively preserves high-level semantic information and fine-grained spatial structures without relying on the underlying backbone network.

\subsection{Ultra-lightweight Performance}
Tab. \ref{tab:efficiency} presents a comprehensive evaluation of computational efficiency, memory footprint, and inference speed for RaysUp in comparison with existing upsampling methods across multiple resolutions. RaysUp demonstrates a clear advantage in all metrics. At a standard input resolution of $224 \times 224$, RaysUp requires only 0.14M parameters, consumes 10.17 GFLOPs, and uses 1.26GB of GPU memory, achieving 55 FPS, substantially outperforming alternatives such as FeatUp \cite{fu2024featup}, LoftUp \cite{iccv2025loftup}, JAFAR \cite{couairon2025jafar}, and AnyUp \cite{wimmer2026anyup}. As the target resolution scales to $448 \times 448$ and $896 \times 896$, RaysUp maintains a low computational burden while delivering high inference speeds (27 and 8 FPS, respectively), whereas other methods experience dramatic slowdowns or run out of memory (OOM) at higher resolutions. Even at extreme resolutions of $2K \times 2K$, RaysUp remains runnable (1 FPS), highlighting its scalability.

These results confirm that RaysUp achieves an ultra-lightweight design that balances parameter efficiency, computational cost, and high-speed inference, making it suitable for dense prediction tasks across a wide range of resolutions while maintaining practical usability on modern GPUs.

\begin{table*}[!t]
    \centering
   \caption{\textbf{Ultra-lightweight Performance.} Comparison of Parameters (M), GFLOPs, GPU memory (GB), and inference FPS at various upsampling resolutions.}
    \resizebox{\linewidth}{!}{
    \begin{tabular}{l c | c c c | c c c | c c c | c}
        \toprule
        & & \multicolumn{3}{c|}{$224 \times 224$} & \multicolumn{3}{c|}{$448 \times 448$} & \multicolumn{3}{c|}{$896 \times 896$} & \multicolumn{1}{c}{$2K \times 2K$} \\
        \cmidrule(lr){3-5} \cmidrule(lr){6-8} \cmidrule(lr){9-11} \cmidrule(lr){12-12}
        Method & Params (M) & GFLOPs $\downarrow$ & Mem $\downarrow$ & FPS $\uparrow$ & GFLOPs $\downarrow$ & Mem $\downarrow$ & FPS $\uparrow$ & GFLOPs $\downarrow$ & Mem $\downarrow$ & FPS $\uparrow$ & FPS $\uparrow$ \\
        \midrule
        FeatUp \cite{fu2024featup}  & 0.17 & 31.79 & 1.73 & 48 & 127.02 & 4.08 & 14 & 507.95 & 15.49 & 2 & OOM \\
        LoftUp \cite{iccv2025loftup}  & 4.30 & 397.32 & 2.88 & 7 & 1970.70 & 6.39 & 4 & OOM & OOM & OOM & OOM \\
        JAFAR \cite{couairon2025jafar}   & 0.62 & 91.88 & 2.01 & 16 & 366.52 & 5.98 & 11 & 1465.08 & 21.86 & 2 & OOM \\
        AnyUp \cite{wimmer2026anyup}   & 0.87 & 84.21 & 2.19 & 11 & 328.55 & 6.73 & 5 & 1306.27 & 24.85 & 1 & OOM \\
        \textbf{RaysUp} & \textbf{0.14} & \textbf{10.17} & \textbf{1.26} & \textbf{55} & \textbf{40.67} & \textbf{2.69} & \textbf{27} & \textbf{162.68} & \textbf{8.47} & \textbf{8} & \textbf{1} \\
        \bottomrule
    \end{tabular}
    }
    \label{tab:efficiency}
    \vspace{-15px}
\end{table*}

\begin{table*}[ht!]
    \centering
    \caption{\textbf{Ablation Study.} Default settings of RaysUp are marked with an asterisk(*).}
    \resizebox{0.85\textwidth}{!}{
    \begin{tabular}{l c c c c c | c}
        \hline
        \rule[-1.5ex]{0pt}{4ex} & {DINOv2} & {SigLIP2} & {PE Spatial} & {DINOv3} & Avg. & Params (M) \\
        \hline
        
        \textbf{Guidance Encoder} & & & & & & \\
        \quad Single-Branch     & 84.37 & 79.17 & 77.34 & 86.63 & 81.87 & 0.266 \\
        \quad Dual-Branch       & 84.49 & 79.25 & 77.39 & 87.02 & 82.03 & 0.66 \\
        \quad Multi-Branch      & 84.52 & 79.33 & 77.42 & 87.11 & 82.09 & 0.268 \\
        \quad Decoupled-Branch * & \textbf{84.64} & \textbf{79.37} & \textbf{77.47} & \textbf{87.20} & \textbf{82.17} & \textbf{0.14} \\
        \hline

        \textbf{Guidance Feature Dim} & & & & & & \\
        \quad $D_g = 128$       & 84.26 & \textbf{79.72} & 77.35 & 86.78 & 82.02 & \textbf{0.06} \\
        \quad $D_g = 256$ * & \textbf{84.64} & 79.37 & 77.47 & 87.20 & 82.17 & 0.14 \\
        \quad $D_g = 512$       & 84.61 & 79.46 & 77.49 & 87.24 & \textbf{82.20} & 0.40 \\
        \quad $D_g = 768$       & 84.59 & 79.41 & \textbf{77.50} & \textbf{87.26} & 82.19 & 0.89 \\
        \hline
        
        \textbf{Guidance Conv Blocks} & & & & & & \\
        \quad $L = 1$ *       & 84.64 & 79.37 & 77.47 & 87.20 & 82.17 & \textbf{0.14} \\
        \quad $L = 2$  & \textbf{84.66} & 79.59 & 77.53 & 87.42 & 82.30 & 0.19 \\
        \quad $L = 3$         & 84.64 & \textbf{79.68} & 77.95 & 87.51 & \textbf{82.44} & 0.27 \\
        \quad $L = 4$          & 84.40 & 79.60 & \textbf{78.08} & \textbf{87.66} & 82.43 & 0.36 \\
        \hline
        \hline
      
        \textbf{Positional Encoding} & & & & & & \\
        \quad None            & 83.51 & 77.70 & 76.47 & 86.54 & 81.05 & 0.14 \\
        \quad RoPE \cite{su2024rope} & 84.43 & 78.89 & 77.24 & 87.15 & 81.92 & 0.14 \\
        \quad SinRays \cite{icml2024sinrays} & 84.12 & 78.45 & 76.81 & 86.98& 81.59 & 0.47 \\
        \quad RayPE * & \textbf{84.64} & \textbf{79.37} & \textbf{77.47} & \textbf{87.20} & \textbf{82.17} & \textbf{0.14} \\
        \hline
          
        \textbf{Image Pose} ($\mat{T}$) & & & & & & \\
        \quad Identity *      & 84.64 & 79.37 & 77.47 & 87.20 & 82.17 & \textbf{0.14} \\
        \quad DA3-Small \cite{lin2026da3}     & 84.69 & \textbf{79.46} & 77.85 & 87.65 & 82.41 & 0.14 \\
        \quad DA3-Base \cite{lin2026da3}     & \textbf{84.87} & 79.21 & 77.90 & 87.78 & \textbf{82.44} & 0.14 \\
        \quad DA3-Large  \cite{lin2026da3}    & 84.76 & 79.06 & \textbf{78.03} & \textbf{87.84} & 82.42 & 0.14 \\
        \hline
    \end{tabular}}
    \label{tab:ablation}
    \vspace{-15px}
\end{table*}

\subsection{Ablation Study}
We conducted a series of ablation experiments to evaluate the effectiveness of the core components within the RaysUp framework. All experiments were systematically performed across four representative VFMs, including DINOv2 \cite{Oquab24dinov2}, SigLIP2 \cite{tschannen2025siglip2}, PE Spatial \cite{bolya2025pespatial}, and DINOv3 \cite{simeoni2025dinov3}.

\textbf{Effectiveness of Spatially Decoupled Guidance Encoder.}
As shown in Tab. \ref{tab:ablation}, compared to Single-Branch, Dual-Branch, and Multi-Branch designs, the proposed Decoupled-Branch achieves the highest performance across all four VFMs, with an average of 82.17\%, while drastically reducing the parameter count to only 0.14M. This indicates that decoupled modeling of spatial guidance features not only more effectively captures multi-scale spatial semantic information but also enables an ultra-lightweight design, substantially improving efficiency for downstream tasks. Regarding the guidance feature dimension and the number of convolutional blocks, performance varies across different VFMs; considering the trade-off between accuracy and efficiency, RaysUp adopts a guidance feature dimension of $D_g = 256$ with a single convolutional block.

\textbf{Effectiveness of RayPE.}
As shown in Tab. \ref{tab:ablation}, omitting positional encoding leads to a significant drop in performance (average 81.05\%). While existing methods such as RoPE \cite{su2024rope} and SinRays \cite{icml2024sinrays} provide moderate improvements, RayPE achieves the highest performance across all VFMs without introducing additional parameters. This indicates that RayPE effectively injects implicit 3D geometric priors, enhancing geometric consistency, boundary fidelity, and structural reconstruction. Moreover, incorporating DA3 \cite{lin2026da3} pose information can further improve performance in certain scenarios; however, considering computational efficiency, RaysUp adopts an Identity pose configuration.

\section{Conclusion}
We introduced RaysUp, an ultra-lightweight, task-agnostic, and encoder-agnostic framework for universal feature upsampling, capable of reconstructing backbone features at arbitrary resolutions. By integrating a Spatially Decoupled Guidance Encoder, an Any-Resolution Cross-Attention mechanism, and Ray Positional Encoding (RayPE), RaysUp injects implicit 3D geometric priors to achieve geometry-aware high-resolution feature reconstruction. Extensive experiments demonstrate that RaysUp delivers superior performance across multiple dense prediction tasks, using only about $1/5$ of the parameters of baseline methods and achieving approximately 7$\times$ faster inference, while effectively balancing accuracy and computational efficiency.

\section*{Acknowledgements}

This work was supported in part by the New Generation Artificial Intelligence-National Science and Technology Major Project under Grant 2025ZD0123701, in part by the National Natural Science Foundation of China under Grant 62476202 and 62272343, and in part by the Fundamental Research Funds for the Central Universities.

\bibliographystyle{splncs04}
\bibliography{main}

\clearpage
\appendix

\begin{center}
    {\Large RaysUp: Ultra-light Universal Feature Upsampling via Geometry-Aware Ray Representation} \\[0.3em]
    {\large --- Supplementary Material ---}
\end{center}

\section{Motivation of Spatially Decoupled Guidance Encoder}
\label{sec:motivation_sdge}
\begin{figure*}
    \centering
    \includegraphics[width=\linewidth]{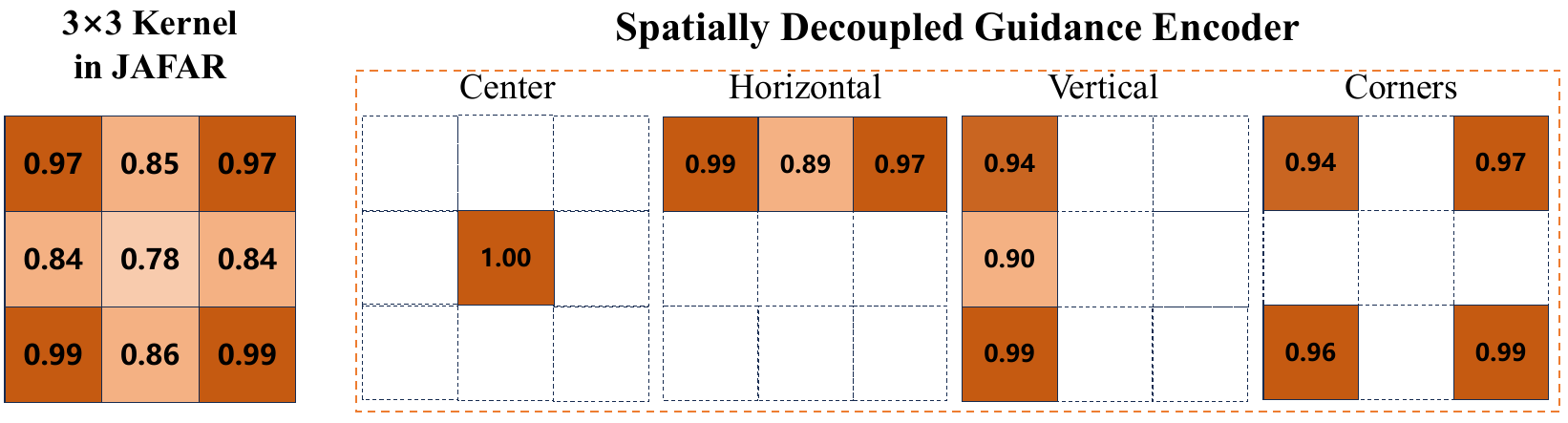}
   \caption{\textbf{Visualization of convolutional kernel weights.} Conventional kernels in JAFAR \cite{couairon2025jafar} assign lower weights to the center and higher weights to corners and edges, potentially reducing central channel mixing and causing holes in upsampled features. In contrast, the Spatially Decoupled Guidance Encoder increases central weights to 1.00 and employs independent horizontal, vertical, and corner branches, enhancing spatial consistency and preserving semantic continuity during cross-attention.}
    \label{fig:conv}
\end{figure*}

As demonstrated by ACNet \cite{ding2019acnet} and RepVGG \cite{ding2021repvgg}, trained convolutional kernels often exhibit sparsity patterns with distinct structural characteristics. In semantic-continuous visual foundation models, uneven spatial distributions of image-guided features may lead to discontinuities in upsampled backbone features. To investigate this phenomenon, as shown in Fig. \ref{fig:conv}, we visualized the weight distribution of $3\times 3$ convolutional kernels in JAFAR \cite{couairon2025jafar}. The results indicate that conventional $3\times 3$ convolutions assign relatively low weights to the center ($0.78$), while corners and certain edges receive higher weights ($0.97-0.99$). This imbalance may hinder effective channel mixing at the center pixels, potentially explaining the appearance of holes in the upsampled features and semantic segmentation outputs in JAFAR (see Fig. \ref{fig:task_baseline}).

In contrast, our Spatially Decoupled Guidance Encoder effectively addresses this limitation: the central weights are substantially increased to the normalized maximum of 1.00, facilitating channel interactions analogous to a $1\times1$ convolution. Simultaneously, independent horizontal ($1\times3$), vertical ($3\times 1$), and corner convolutional branches maintain high spatial sensitivity in peripheral regions ($0.89-0.99$), enhancing the overall spatial consistency of the features. This continuous and uniform feature structure ensures that, during cross-attention, high-resolution guidance query features $\mathcal{Q}_{ray}$ can interact with low-resolution key features $\mathcal{K}_{ray}$ and visual foundation model features $\mathcal{V}$ while preserving spatial semantic continuity.

\section{Implicit Geometric Injection in RayPE}
\begin{figure}[t]
  \centering
  \includegraphics[width=\linewidth]{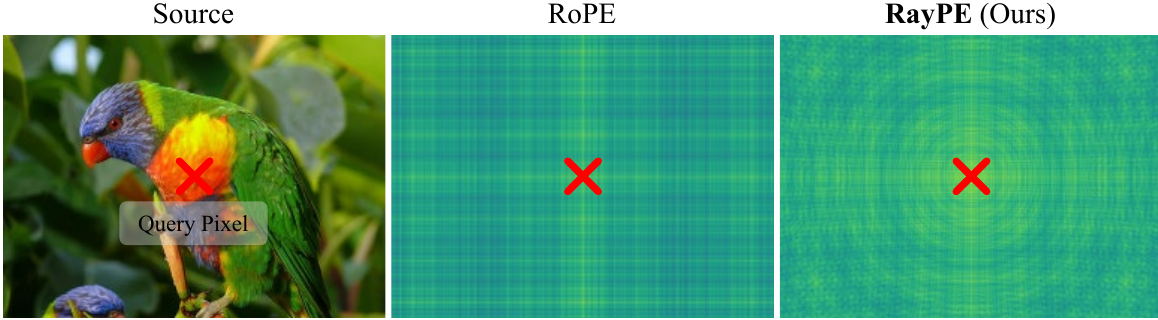}
  \caption{Attention bias map from the query pixel (\textcolor{red}{×}).}
  \label{fig:rope_vs_raype}
\end{figure}

RoPE is defined on a planar isotropic 2D Euclidean grid, where the positional phase is linearly associated with pixel coordinates $(i,j)$: $\boldsymbol{\theta}_{\mathrm{RoPE}}(i,j)=[i\omega,\; j\omega]^T$ ($\omega$ denotes the frequency). In contrast, RayPE implicitly encodes 3D spatial positions using normalized camera rays. \textit{Assuming an identity camera extrinsic matrix}, the positional phase is:
\begin{equation*}
    \scriptsize
    \boldsymbol{\theta}_{\mathrm{RayPE}}(i,j) = \omega \cdot \frac{[i,j,f]^T}{\sqrt{i^2+j^2+f^2}},
\end{equation*}
where $f$ denotes the focal length. During upsampling, the positional phase in RoPE shifts with increasing coordinates, requiring additional interpolation for alignment. In contrast, RayPE preserves geometric feature consistency under identical viewing directions, thereby naturally achieving scale equivariance. As shown in Fig. \ref{fig:rope_vs_raype}, RayPE extends positional encoding from a 2D image grid to the unit viewing sphere, enabling the attention mechanism to model angular consistency rather than pixel-distance consistency.

\section{Further Implementation Details}
\subsection{Hyperparameter Settings}
The key hyperparameters of our model were set as follows. The dimensionality of the guidance features was $D_g = 256$, and the total number of frequency bands used in Ray Positional Encoding was $N = 5$. For the ray origin, the maximum and minimum spatial wavelengths were set to $\lambda_{\max} = 4.0$ and $\lambda_{\min} = 4.0/81$, respectively, determining the lowest and highest angular frequencies. For the ray direction, the maximum and minimum spatial wavelengths were set to $\lambda_{\max} = 8.0$ and $\lambda_{\min} = 8.0/81$, respectively. In the cross-attention mechanism, the feature dimensionality of each attention head was $d_{\text{head}} = 4$. Within the Neighborhood Cross-Attention module, a $k \times k$ local neighborhood centered at each correspondence was considered, with $k = 6$.

\subsection{Task Settings}
\subsubsection{Semantic Segmentation.}
In our semantic segmentation experiments, the pre-trained parameters were retained, and linear classifiers were trained exclusively on the extracted features. Models were optimized using a cross-entropy loss, with both input and target images uniformly resized to $448 \times 448$. Parameter updates were performed using the AdamW \cite{adamw2019} optimizer in conjunction with a cosine annealing learning rate schedule, maintaining a fixed learning rate of $5 \times 10^{-4}$ across all datasets. Each model was trained for 20 epochs on the Pascal VOC \cite{ijcv2015voc} (21 classes), ADE20K \cite{zhou2019ade20k} (151 classes), Cityscapes \cite{Cordts2016cityscapes} (19 classes), and COCO \cite{caesar2018coco} (27 classes) benchmarks.

\subsubsection{Depth and Surface Normal Estimation.}
We conducted monocular depth and surface normal estimation on the NYUv2 \cite{eccv2012nyuv2} dataset. For the depth range of 0 to 10 meters, the AdaBins \cite{AdaBins2020} model was applied with 256 bins, and optimization was performed by combining scale-invariant sigmoid loss with gradient matching loss. For the surface normal task, the model predicted orthogonal unit vectors and output an associated uncertainty scalar, utilizing the robust loss function proposed by Bae \textit{et al.} \cite{Bae2021}. For both tasks, multi-scale features extracted from three stages of the backbone network were first upsampled before being input to the DPT decoder \cite{depth2021}. The terminal network produced a 256-dimensional output for the depth probability distribution and a 4-dimensional output for the surface normals. During training, both models were configured with the AdamW optimizer \cite{adamw2019} (initial learning rate $0.001$, weight decay $0.01$) and trained for 10 epochs.

\subsubsection{Video object segmentation.}
For the semi-supervised video object segmentation task, the models were evaluated on the DAVIS 2017 validation set \cite{pont2017davis} at 480p resolution. Dense features were extracted with an upsampling factor of 2. During mask prediction, a non-parametric label propagation strategy was employed, in which target features were matched against a memory queue containing the initial frame and the seven preceding frames. To improve robustness, feature matching was restricted to a local spatial neighborhood of 12 pixels, retaining only the top five affinity values for each target pixel. The final segmentation masks were generated by multiplying the source masks with the normalized affinity matrix, followed by bilinear upsampling and an argmax operation.

\subsubsection{Open-Vocabulary Segmentation.}
During the open-vocabulary segmentation inference phase, we followed the configuration of ProxyCLIP \cite{lan2024proxyclip} with a key modification: instead of using traditional bilinear interpolation, the extracted features were upsampled to 4$\times$ resolution using our proposed upsampling module. For dense inference, the extracted patch representations were first projected into the aligned vision-language feature space. Corresponding textual representations were constructed by inputting prompts of the form ``a photo of a \{label name\}.'' into the CLIP \cite{clip2021} text encoder. Patch-level classification was then performed by computing the cosine similarity between visual and textual representations within the shared latent space, and each patch was assigned the semantic label corresponding to the text prompt with the highest similarity. Finally, the predicted patch labels were reshaped and upsampled to generate the final segmentation map for the entire image.

\subsection{Baselines}

\subsubsection{FeatUp \cite{fu2024featup}} 
FeatUp proposed two feature upsampling variants: a feedforward Joint Bilateral Upsampling (JBU) network \cite{KopfCLU07JBU} and an image-specific implicit multi-layer perceptron, both supervised using a multi-view consistency loss across transformed low-resolution feature views. As the official repository only provided pre-trained weights for the JBU variant, our evaluation was conducted exclusively on this architecture. Consequently, spatial enhancement was limited to fixed integer scales through sequential cascading of JBU blocks.

\subsubsection{LoftUp \cite{iccv2025loftup}} 
LoftUp employed a coordinate-based cross-attention Transformer, constructing queries from high-resolution images and spatial coordinates, while using low-resolution VFM features as keys and values. Training involved an initial feature refinement phase supervised by SAM-generated \cite{kirillov2023sam} class-agnostic masks, followed by a self-distillation phase in which an EMA-updated teacher model guided the student network. LoftUp supported only fixed-scale upsampling to match the original input image size. In our evaluation, we utilized the officially released pre-trained weights of LoftUp.

\subsubsection{JAFAR \cite{couairon2025jafar}} 
JAFAR is a task-agnostic feature upsampler that employed a cross-attention mechanism to restore low-resolution features to arbitrary target resolutions. The framework derived asymmetric queries and keys from a shared high-resolution guidance image, with keys further modulated via Spatial Feature Transform (SFT) \cite{sft2018} to incorporate semantic context. During training, the model was optimized using a combination of cosine and L2 losses applied to multi-resolution image views at low upsampling factors. For evaluation, we directly employed the officially released pre-trained weights of JAFAR.

\subsubsection{AnyUp \cite{wimmer2026anyup}} 
Similar to JAFAR \cite{couairon2025jafar}, AnyUp is an attention-based feature upsampler that utilizes a feature-agnostic convolutional layer to map inputs of varying dimensionality into a canonical format by independently convolving and averaging all channels. The architecture implements a local window attention mechanism computed relative to the feature map size, rather than standard global attention. During training, the model employed a crop-based strategy, learning to match upsampled features from a downsampled image to target features extracted from randomly sampled high-resolution local crops. The optimization objective combined cosine similarity and L2 loss, supplemented by self-consistency and input-consistency regularization. This training procedure required approximately 5 hours. In our evaluation, we directly employed the officially released pre-trained weights of AnyUp.

\section{Additional Ablation Study}
\subsection{Ablation Study Implementation Details}

\subsubsection{Guidance Encoder.}
For the single-branch encoder, we adopted the architecture used in AnyUp \cite{wimmer2026anyup}, consisting of a $1 \times 1$ convolution. The dual-branch encoder incorporated parallel $1 \times 1$ and $3 \times 3$ convolutions, with their outputs concatenated along the channel dimension. The multi-branch encoder extended this design further by employing four convolutional branches ($1 \times 1$, $1 \times 3$, $3 \times 1$, and $3 \times 3$), whose outputs were similarly concatenated channel-wise. In all variants, each basic encoder block comprised two consecutive GroupNorm \cite{wu2018groupnorm}, SiLU \cite{elfwing2018silu}, and convolution operations.

\subsubsection{Guidance Feature Dimensionality.} 
We investigated the impact of guidance feature dimensionality in our ablation studies. The results indicated that the 768-dimensional setting underperformed relative to the 512-dimensional configuration on downstream tasks, and neither setting yielded a notable improvement over the 256-dimensional configuration. In contrast, the 256-dimensional representation consistently and substantially outperformed the 128-dimensional baseline. These findings suggest that 256 dimensions provide sufficient representational capacity for the four branches of our decoupled encoder to learn robust and effective guidance features.

\subsubsection{Positional Encoding.} 
In the ablation study on positional encoding, RoPE \cite{su2024rope} was implemented following the protocol established by JAFAR \cite{couairon2025jafar} and AnyUp \cite{wimmer2026anyup}. For SinRays \cite{icml2024sinrays}, ray origins and ray directions were treated as independent components and concatenated with features extracted from the image encoder. A sine activation was subsequently applied to capture high-frequency details, and the resulting representation was projected to the target hidden dimensionality via a convolutional layer.

\subsubsection{Image Pose.} 
Under the identity-pose setting, the extrinsic matrix was set to the identity matrix, while the intrinsic parameters followed a simplified pinhole camera model. The training durations for DA3-Small, DA3-Base, and DA3-Large were approximately 5, 9, and 55 hours, respectively.

\begin{table*}[t]
    \centering
    \caption{\textbf{Ablation Study on Training Strategy.} The experiments were conducted on semantic segmentation tasks.}
    \label{tab:semantic_segmentation}
    \resizebox{\linewidth}{!}{
    \begin{tabular}{l | c c | c c | c c | c c}
        \toprule
        & \multicolumn{2}{c|}{COCO\cite{caesar2018coco}} & \multicolumn{2}{c|}{PASCAL-VOC\cite{ijcv2015voc}} & \multicolumn{2}{c|}{ADE20k\cite{zhou2019ade20k}} & \multicolumn{2}{c}{Cityscape\cite{Cordts2016cityscapes}} \\
        & mIoU ($\uparrow$) & Acc($\uparrow$) & mIoU ($\uparrow$) & Acc($\uparrow$) & mIoU ($\uparrow$) & Acc ($\uparrow$) & mIoU ($\uparrow$) & Acc($\uparrow$) \\
        \midrule
        Bilinear    & 59.58 & 79.42 & 81.70 & 95.44 & 40.47 & 74.13 & 59.72 & 92.55 \\
        FeatUp\cite{fu2024featup} & 61.89 & 81.10 & 83.37 & 96.01 & 42.33 & 75.65 & 60.18 & 93.05 \\
        LoftUp\cite{iccv2025loftup} & 62.23 & 81.38 & 84.50 & 96.30 & 42.17 & 75.79 & \underline{62.09} & 93.54 \\
        JAFAR\cite{couairon2025jafar}   & 61.79 & 81.11 & 83.89 & 96.11 & 42.16 & 75.56 & 61.40 & 93.47 \\
        AnyUp\cite{wimmer2026anyup}  & 62.14 & 81.38 & 84.18 & 96.20 & 42.15 & 75.71 & 60.62 & 93.26 \\
        \textbf{RaysUp} (w/o Crop) & \underline{62.32} & \underline{81.47} & \underline{84.64} & \underline{96.34} & \underline{42.34} & \underline{75.81} & 61.88 & \underline{93.64} \\
        \textbf{RaysUp} (w/ Crop)     & \textbf{62.47} & \textbf{81.55} & \textbf{84.74} & \textbf{96.38} & \textbf{42.60} & \textbf{75.87} & \textbf{63.04} & \textbf{93.95} \\
        \bottomrule
    \end{tabular}
    }
\end{table*}

\subsection{Ablation Study on Training Strategy} 

Following the training paradigm of existing feature upsampling methods \cite{wimmer2026anyup}, RaysUp also employed a lightweight training strategy based on local image crops. Specifically, given a high-resolution image $\ten{I}_{hr} \in \mathbb{R}^{3 \times H_{in} \times W_{in}}$, a smaller local image crop $\ten{I}'_{hr} \in \mathbb{R}^{3 \times H_{crop} \times W_{crop}}$ was randomly sampled spatially. Simultaneously, the original image $\ten{I}_{hr}$ was downsampled to the same spatial dimensions as $\ten{I}'_{hr}$, yielding a low-resolution image $\ten{I}_{lr} \in \mathbb{R}^{3 \times H_{crop} \times W_{crop}}$. Features were then extracted from both images using a frozen vision encoder: the low-resolution source features $\ten{F}^{lr} = e(\ten{I}_{lr}) \in \mathbb{R}^{D_f \times H_{lr} \times W_{lr}}$ were obtained from the downsampled image, while the high-resolution target features $\ten{F}_{tgt} = e(\ten{I}'_{hr}) \in \mathbb{R}^{D_f \times H_{lr} \times W_{lr}}$ were extracted from the local crop to serve as the supervisory signal.

Subsequently, RaysUp upsampled $\ten{F}^{lr}$ to reconstruct the full high-resolution feature map $\hat{\ten{F}}^{hr} = f(\ten{I}_{hr}, \ten{F}^{lr}) \in \mathbb{R}^{D_f \times H_{hr} \times W_{hr}}$. To ensure precise local alignment, a feature sub-region $\hat{\ten{F}}^{hr}_{crop} \in \mathbb{R}^{D_f \times H_{lr} \times W_{lr}}$ was cropped from $\hat{\ten{F}}^{hr}$ to correspond exactly to the spatial location of $\ten{I}'_{hr}$, matching the scale of $\ten{F}_{tgt}$. The reconstruction loss was then computed by jointly measuring the cosine similarity and $L_2$ distance between these local feature regions:
\begin{equation}
    \mathcal{L}(\hat{\ten{F}}^{hr}_{crop}, \ten{F}_{tgt}) = 1 - \cos(\hat{\ten{F}}^{hr}_{crop}, \ten{F}_{tgt}) + \|\hat{\ten{F}}^{hr}_{crop} - \ten{F}_{tgt}\|_2^2.
\end{equation}
With approximately 4 hours of training, the results in \cref{tab:semantic_segmentation} demonstrated consistent performance improvements, further validating the effectiveness and scalability of the proposed model.

\begin{table*}[t]
    \centering
    \caption{\textbf{Upsampling from Any to Any Resolution.}  The experiments were conducted on  Pascal-VOC \cite{ijcv2015voc} dataset.}
    \label{tab:anytoany}
    \resizebox{\linewidth}{!}{
    \begin{tabular}{l | c c | c c | c c | c c | c c | c c}
        \toprule
        & \multicolumn{2}{c|}{$16 \rightarrow 112$} & \multicolumn{2}{c|}{$16 \rightarrow 448$} & \multicolumn{2}{c|}{$16 \rightarrow 896$} & \multicolumn{2}{c|}{$32 \rightarrow 112$} & \multicolumn{2}{c|}{$32 \rightarrow 224$} & \multicolumn{2}{c}{$32 \rightarrow 896$} \\
        & mIoU ($\uparrow$) & Acc ($\uparrow$) & mIoU ($\uparrow$) & Acc ($\uparrow$) & mIoU ($\uparrow$) & Acc ($\uparrow$) & mIoU ($\uparrow$) & Acc ($\uparrow$) & mIoU ($\uparrow$) & Acc ($\uparrow$) & mIoU ($\uparrow$) & Acc ($\uparrow$) \\
        \midrule
        Bilinear    & 75.91 & 93.71 & 75.76 & 93.69 & 75.66 & 93.65 & 81.56 & 95.40 & 81.42 & 95.38 & 81.71 & 95.43 \\
        FeatUp \cite{fu2024featup} & 78.13 & 94.50 & 78.24 & 94.54 & 78.25 & 94.54 & 83.23 & 95.97 & 83.36 & 96.01 & 83.37 & 96.01 \\
        LoftUp \cite{iccv2025loftup} & 79.87 & 95.05 & 79.68 & 94.88 & 79.56 & 94.83 & 84.41 & 96.28 & \underline{84.64} & \underline{96.32} & \underline{84.45} & \underline{96.30} \\
        JAFAR \cite{couairon2025jafar}   & \textbf{81.35} & \textbf{95.43} & \textbf{80.85} & \textbf{95.30} & \textbf{80.81} & \textbf{95.28} & 84.49 & 96.27 & 84.40 & 96.25 & OOM & OOM \\
        AnyUp\cite{wimmer2026anyup}  & 80.00 & 95.10 & 79.79 & 95.04 & OOM & OOM & \underline{84.64} & \underline{96.31} & 84.56 & 96.31 & 84.08 & 96.16 \\
        \textbf{RaysUp} & \underline{80.44} & \underline{95.17} & \underline{80.37} & \underline{95.18} & \underline{80.33} & \underline{95.16} & \textbf{84.76} & \textbf{96.35} & \textbf{84.81} & \textbf{96.37} & \textbf{84.67} & \textbf{96.35} \\
        \bottomrule
    \end{tabular}
    }
\end{table*}

\subsection{Upsampling from Any to Any Resolution.}
Following our semantic segmentation configurations, we evaluated the generalization capability of each model for upsampling arbitrary input resolutions to arbitrary target resolutions on the Pascal-VOC \cite{ijcv2015voc} dataset. Since the JBU \cite{KopfCLU07JBU} variant of FeatUp \cite{fu2024featup} is restricted to a fixed $16\times$ upsampling factor through stacked JBU blocks, and LoftUp \cite{iccv2025loftup} only supports upsampling to the original image resolution, we applied bilinear interpolation to align their outputs with the target resolutions. 

The quantitative results are summarized in \cref{tab:anytoany}. As observed, our proposed RaysUp consistently achieved the best or second-best performance across a wide range of resolution pairs. Notably, JAFAR \cite{couairon2025jafar}, although theoretically supporting any-to-any upsampling, demonstrated competitive performance primarily at lower resolutions (e.g., $16 \rightarrow 112$ and $16 \rightarrow 448$). At higher resolutions, the model encountered memory limitations (OOM), preventing evaluation at the largest scales. In contrast, RaysUp maintained strong performance across all resolutions, consistently surpassing AnyUp \cite{wimmer2026anyup}, a comparable method designed to handle arbitrary Vision Foundation Models (VFMs) and arbitrary output resolutions. These results highlight the scalability and robustness of RaysUp in practical high-resolution scenarios, demonstrating its ability to generalize effectively to unseen resolution combinations while maintaining superior semantic accuracy.

\section{Further Qualitative Experiments}

\begin{figure*}
    \centering
    \includegraphics[width=\linewidth]{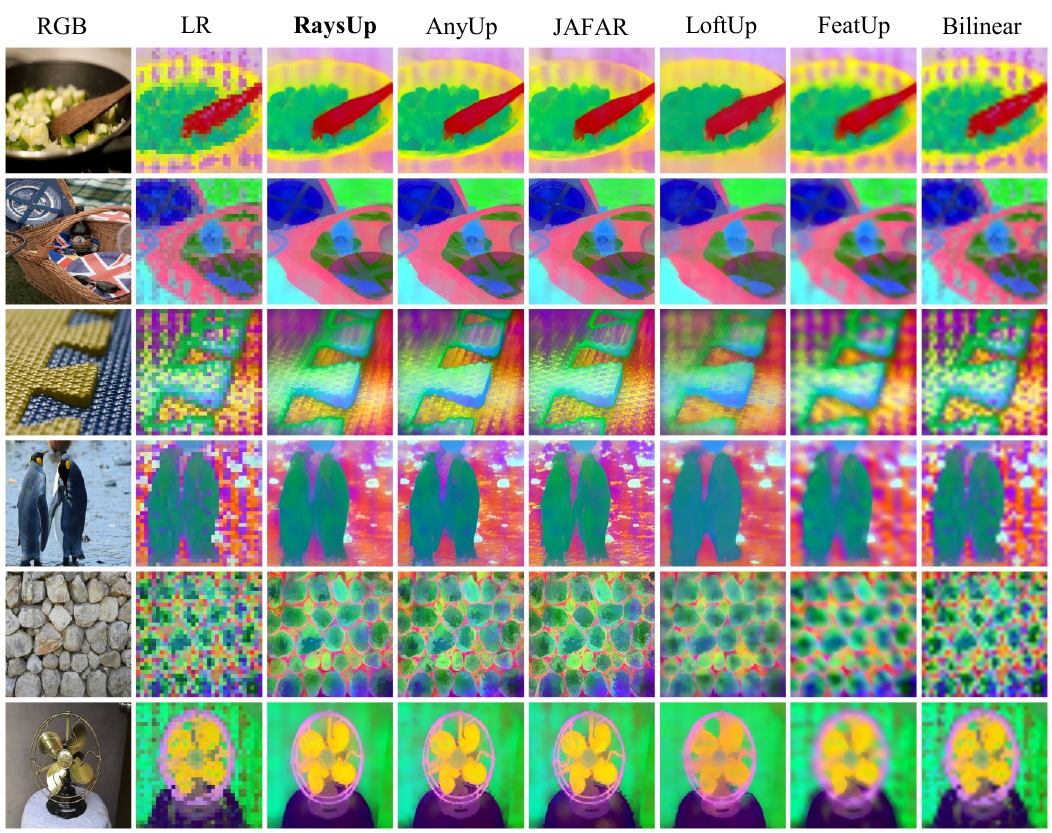}
   \caption{\textbf{Additional pca visualizations on ImageNet.} For consistent visualization, a uniform PCA basis is utilized across all methods. The results demonstrate that only RaysUp, AnyUp, and JAFAR yield distinctly sharp PCA projections, successfully maintaining the underlying feature space during upsampling.}
    \label{fig:pca}
\end{figure*}

\begin{figure*}
    \centering
    \includegraphics[width=\linewidth]{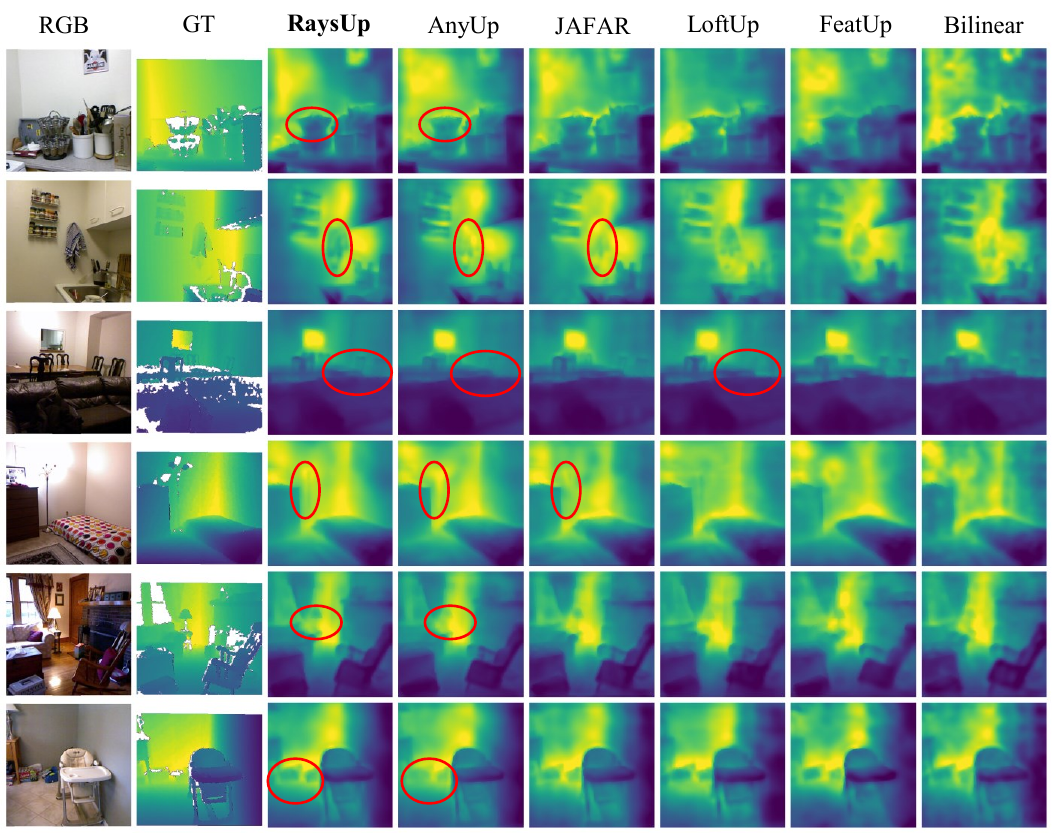}
   \caption{\textbf{Additional depth estimation visualizations on NYUv2.} Compared with baseline methods, RaysUp preserves sharper object boundaries and produces predictions that are more closely aligned with the ground truth.}
    \label{fig:depth}
\end{figure*}

\begin{figure*}
    \centering
    \includegraphics[width=\linewidth]{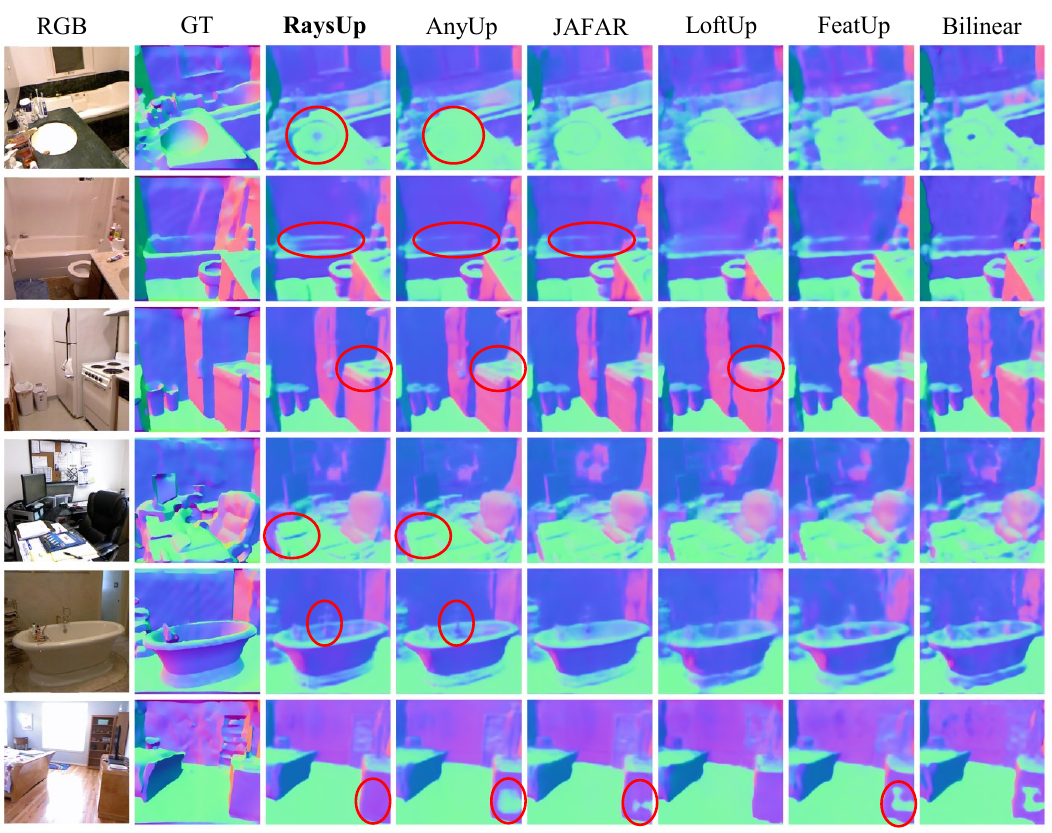}
   \caption{\textbf{Additional surface normal estimation visualizations on NYUv2.} In some cases (e.g., rows 2 and 5), RaysUp produces predictions that are visually more coherent than the provided ground truth, whereas competing methods tend to introduce noticeable artifacts in challenging regions (e.g., row 6).}
    \label{fig:norm}
\end{figure*}

\begin{figure*}
    \centering
    \includegraphics[width=\linewidth]{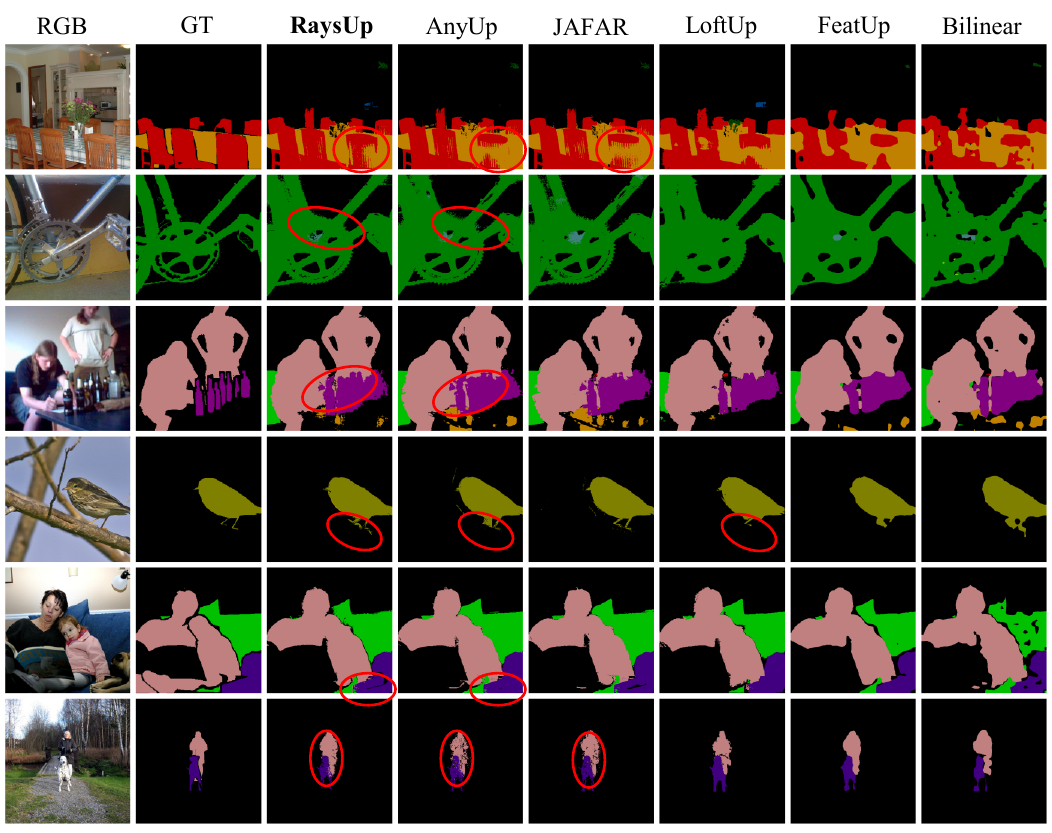}
   \caption{\textbf{Additional semantic segmentation visualizations on VOC.} Compared to AnyUp, RaysUp yields more complete segmentation masks with sharper boundaries, as observed in the third and sixth rows. Furthermore, in certain instances, our results exhibit finer details than even the ground truth annotations (e.g., the second row).}
    \label{fig:seg}
\end{figure*}

\begin{figure*}
    \centering
    \includegraphics[width=\linewidth]{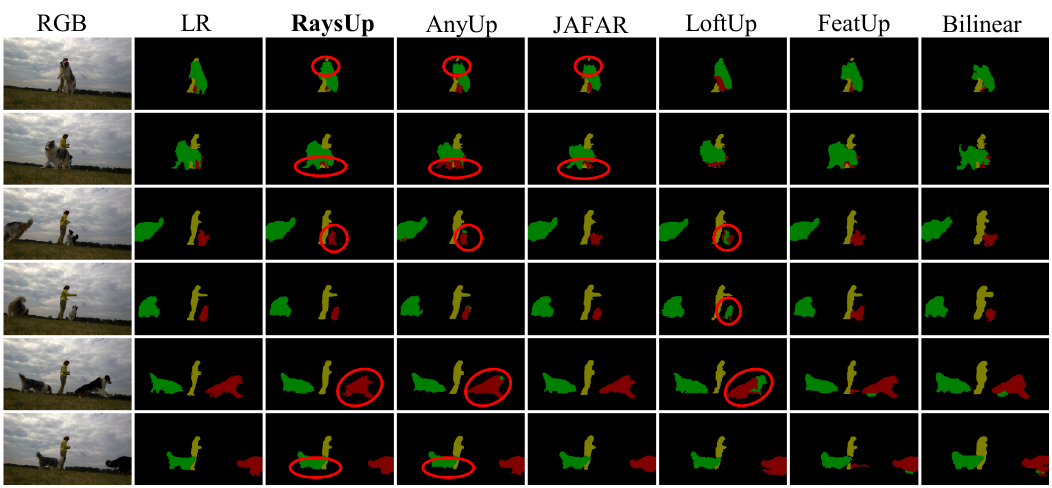}
   \caption{\textbf{Additional video semantic segmentation visualizations on the DAVIS dataset.} We provide qualitative segmentation results for the $8^{\text{th}}$, $11^{\text{th}}$, $24^{\text{th}}$, $28^{\text{th}}$, $37^{\text{th}}$, and $43^{\text{rd}}$ frames of the \textit{dogs-jump} video sequence. As illustrated, RaySup demonstrates superior performance in video semantic segmentation compared to existing methods. Specifically, our approach yields significantly sharper object boundaries and maintains highly robust inter-frame temporal consistency.}
    \label{fig:seg}
\end{figure*}

\end{document}